\documentclass[11pt]{article}
\usepackage[margin=1in]{geometry}
\usepackage[T1]{fontenc}
\usepackage[utf8]{inputenc}
\usepackage{amsmath,amssymb}
\usepackage{booktabs}
\usepackage{graphicx}
\usepackage{xcolor}
\usepackage{enumitem}
\usepackage{environ}
\usepackage{etoolbox}
\usepackage[super,comma,sort&compress]{natbib}
\usepackage[colorlinks=true,linkcolor=blue,citecolor=blue,urlcolor=blue]{hyperref}

\newcommand{\pneg}{p(\textsc{negative})}
\newcommand{\dauc}{\Delta\mathrm{AUROC}}
\renewcommand{\figurename}{Fig.}
\makeatletter

\newif\ifEDnext
\newcommand{\EDtable}{\global\EDnexttrue}
\newcommand{\EDqueue}{}

\RenewEnviron{table}[1][tbp]{%
  \ifEDnext
    \global\EDnextfalse
    \xappto\EDqueue{\noexpand\EDprint{\unexpanded\expandafter{\BODY}}}%
  \else
    \begin{EDfloat}[#1]\BODY\end{EDfloat}%
  \fi}
\newcommand{\EDprint}[1]{\begin{EDfloat}[!htbp]\EDsetup#1\end{EDfloat}}
\newcommand{\EDsetup}{\renewcommand{\tablename}{Extended Data Table}\let\c@table\c@EDtab
  \renewcommand{\theHtable}{ED.\arabic{table}}}
\newcommand{\printEDtables}{\EDqueue\clearpage}
\makeatother

\newcounter{suppnote}
\newcommand{\suppnote}[2]{%
  \refstepcounter{suppnote}%
  \section*{Supplementary Note \thesuppnote: #1}\label{#2}}

\title{Similar patients make language models predict death less often, not better}
\author{
  Pengyu Zhan\textsuperscript{1},
  Shansong Wang\textsuperscript{1},
  Mojtaba Safari\textsuperscript{1},
  Mingzhe Hu\textsuperscript{1},\\
  Suyang Xi\textsuperscript{1},
  Xiaofeng Yang\textsuperscript{1,*}\\[1ex]
  {\small \textsuperscript{1}Department of Radiation and Cellular Oncology, The University of Chicago, Chicago, IL, USA}\\[0.5ex]
  {\small \textsuperscript{*}Corresponding author: Xiaofeng Yang (xfyang@uchicago.edu)}
}
\date{}

\begin{document}
\maketitle

\begin{abstract}
Large language models can estimate prognosis from clinical notes without task-specific training, and showing them the notes and outcomes of similar patients, a strategy known as dynamic few-shot prompting, reportedly improves such predictions. This comparison, however, changes the wording of the prompt, the presence of labelled examples and their similarity to the patient at once, and in-context examples can shift a model's answers through the labels they display. Here we show, in a controlled replication across 11 cancer cohorts (14{,}164 records; 2{,}282 deaths) from a publicly accessible single-centre database, that similar patients did not help a language model rank patients by their risk of death within 90 days of discharge. With the retriever of the original method, similar examples were equivalent within the pre-specified margin to random ones, and both lowered discrimination relative to the discharge summary alone; a stronger retriever and a larger model reduced this loss but did not reverse it. Examples nonetheless raised accuracy at the default threshold because the zero-shot prompt over-predicted death and examples made that prediction less frequent, and a threshold fitted to zero-shot scores raised accuracy further. The model's answers followed the number of deaths among the examples shown. Evaluations of retrieval-based prompting therefore need random-example and wording-matched controls and threshold-free metrics.
\end{abstract}

When a patient with cancer leaves hospital, the oncology team must decide how soon to arrange follow-up and whether conversations about supportive care and prognosis can wait until the next clinic visit. These decisions are consequential because death may precede a planned visit: in one series of 358 patients discharged from an inpatient oncology service to subacute rehabilitation, 74 (21\%) died within 30 days\citep{yeh2019rehabilitation}. Prognostic models built on structured variables such as tumour site, stage, reason for admission, active treatment and age can identify patients with a short life expectancy at discharge\citep{vicenteconesa2022discharge}, yet functional status, symptom burden and response to treatment are often documented only in free text. Text encoders adapted to clinical language predict readmission, mortality and other outcomes from notes\citep{huang2019clinicalbert,vanaken2021clinical,li2023comparative,jiang2023health}, in some cases with the support of retrieved medical literature\citep{naik2022literature}, and generative large language models (LLMs) can be prompted to make such predictions without task-specific training\citep{zhu2024clinicrealm}. Predicting death after discharge from long discharge documents nevertheless remains demanding\citep{yoon2025lcd}.

In-context learning adapts a general-purpose LLM to a clinical task by showing it a small number of worked examples within the prompt while its parameters remain unchanged\citep{brown2020language}. Which examples are shown matters. In general-domain benchmarks, examples retrieved for their semantic similarity to the query outperform randomly chosen ones\citep{liu2022makes}, retrievers can be trained specifically for this purpose\citep{rubin2022learning}, and nearest-neighbour selection of examples is one component of the Medprompt strategy for medical question answering\citep{nori2023medprompt}. In clinical prediction the idea has additional appeal because it mirrors the way clinicians reason from previous similar cases, and retrieval of similar patients' notes has been incorporated into LLM-based risk prediction\citep{datta2026kameleon}. \citet{anil2026dynamic} proposed a dynamic prompting framework for cancer outcome prediction in which each patient's notes are embedded, the most similar patients in a labelled training set are retrieved, and their notes and outcomes are placed before the target patient's notes as few-shot examples. In breast cancer and glioma cohorts from a single institution, this approach improved on zero-shot prompting, most clearly for glioma, for which the area under the receiver operating characteristic curve (AUROC) of the larger of two Llama-3 models rose from 0.742 with full notes alone to 0.813 with five retrieved, summarised examples.

What produces such gains is less clear. A comparison between dynamic few-shot prompting and zero-shot prompting changes at least three things at once: the wording of the instruction, which typically differs between zero-shot and few-shot templates; the presence of labelled examples of any kind; and the similarity of those examples to the target patient. Only the last corresponds to the clinical rationale of learning from similar patients. Model outputs are sensitive to the order in which examples are presented\citep{lu2022fantastically} and to the frequency and recency of the labels they display\citep{zhao2021calibrate}, and in classification tasks replacing the correct labels of the demonstrations with random ones often changes performance only slightly\citep{min2022rethinking}, which suggests that demonstrations can act through the format and label distribution they convey rather than through the pairing of inputs with labels. Prompts that carry several full-length notes are also long, and models use information in long contexts less reliably\citep{liu2024lost}. How benefit is measured matters as well. Accuracy at a fixed cut-off depends on how readily a model gives one answer as well as on how well it ranks patients, so a prompt that makes an over-called answer less frequent can raise accuracy without any improvement in discrimination, a property that calibration methods for in-context learning exploit by moving the decision boundary\citep{zhao2021calibrate,zhou2024batch}. To our knowledge, dynamic prompting for clinical outcome prediction has not been compared with a random-example control or with a zero-shot prompt of matched wording.

Here we report a controlled conceptual replication of dynamic few-shot prompting in 11 cancer cohorts from the publicly accessible MIMIC-IV database. We retained the retrieval encoder, the prompt wording and the scoring procedure of the original study and applied them, with an open-weight LLM, to the prediction of all-cause death within 90 days of hospital discharge from the discharge summary. To these we added three controls: randomly selected examples, a zero-shot prompt whose wording matches that of the few-shot prompt, and similar examples with shuffled outcome labels. The primary question, whether similar examples improve risk discrimination over random examples, was specified together with an equivalence margin before inference on a held-out test set. We further examined whether a decision threshold can reproduce the accuracy gain obtained with examples, how prompting compares with supervised text classifiers and with a linear probe of the model's hidden state\citep{alain2017understanding}, a type of readout used to ask whether models encode more than their outputs express\citep{burns2023discovering} and to predict outcomes from LLM representations of health records\citep{hegselmann2025llmehr}, and how the displayed examples influence the model's answer.

\subsection*{Study population and design}

The study population comprised 14{,}164 patient--cancer records from 13{,}351 adults who were discharged alive after an index admission that carried a diagnosis of one of 11 cancer types (Extended Data Table~\ref{tab:cohort}). All-cause death within 90 days of the prediction time, defined as the later of discharge and availability of the summary, occurred in 2{,}282 records (16.1\%), ranging from 6.5\% in breast cancer to 27.4\% in pancreatic cancer. Discharge summaries were long (median 2{,}512 to 3{,}545 tokens by cancer), and 37--66\% of the notes in each cohort exceeded the 3{,}000-token limit and were shortened. Each cohort was divided once, by patient and with stratification by outcome, into an example pool, a development set and a held-out test set. The test set, on which all results below are based unless stated otherwise, comprised 5{,}666 records from 5{,}514 patients, with 912 deaths (16.1\%).

\EDtable
\begin{table}[!htbp]
\centering
\caption{\textbf{Characteristics of the 11 cancer cohorts.} Counts are patient--cancer records (13{,}351 unique patients). The pool, development (Dev) and test columns give the number of records (deaths) in each partition. Age is the median in years. ``Tokens'' is the median note length in Qwen3-8B tokens and ``$>$3k'' the proportion of notes longer than 3{,}000 tokens, which were shortened. ``Hospice'' is the proportion of patients discharged to hospice.}
\label{tab:cohort}
\small
\resizebox{\linewidth}{!}{%
\begin{tabular}{lrrrrrrrrrr}
\toprule
Cancer type & $N$ & Deaths (\%) & Pool & Dev & Test & Age & Female (\%) & Tokens & $>$3k (\%) & Hospice (\%) \\
\midrule
Lung                 & 2{,}888 & 700 (24.2) & 1{,}433 (347) & 300 (73)  & 1{,}155 (280) & 70 & 52.1 & 3{,}298 & 58.5 & 4.2 \\
Pancreas             & 1{,}385 & 380 (27.4) & 415 (114) & 416 (114) & 554 (152) & 70 & 47.5 & 3{,}094 & 53.6 & 5.8 \\
Liver and intrahepatic bile duct & 1{,}253 & 259 (20.7) & 375 (77) & 376 (78) & 502 (104) & 64 & 28.3 & 2{,}988 & 49.3 & 4.1 \\
Prostate             & 1{,}722 & 148 (8.6)  & 517 (45)  & 516 (44)  & 689 (59)  & 72 & 0.0  & 2{,}834 & 43.6 & 1.2 \\
Colorectal           & 1{,}433 & 148 (10.3) & 430 (45)  & 430 (44)  & 573 (59)  & 67 & 48.6 & 2{,}733 & 41.3 & 2.0 \\
Non-Hodgkin lymphoma & 1{,}695 & 139 (8.2)  & 508 (41)  & 509 (42)  & 678 (56)  & 68 & 44.9 & 3{,}545 & 66.1 & 1.7 \\
Biliary tract        & 441     & 107 (24.3) & 132 (32)  & 132 (32)  & 177 (43)  & 72 & 47.2 & 2{,}946 & 47.8 & 3.6 \\
Breast               & 1{,}550 & 101 (6.5)  & 465 (31)  & 465 (30)  & 620 (40)  & 62 & 98.8 & 2{,}512 & 36.8 & 1.9 \\
Bladder              & 671     & 101 (15.1) & 202 (31)  & 201 (30)  & 268 (40)  & 76 & 22.8 & 2{,}824 & 44.9 & 1.5 \\
Stomach              & 524     & 101 (19.3) & 158 (31)  & 157 (30)  & 209 (40)  & 67 & 34.9 & 3{,}014 & 51.1 & 4.0 \\
Brain (malignant)    & 602     & 98 (16.3)  & 181 (30)  & 180 (29)  & 241 (39)  & 62 & 42.7 & 2{,}752 & 40.9 & 2.3 \\
\midrule
Total                & 14{,}164 & 2{,}282 (16.1) & 4{,}816 (824) & 3{,}682 (546) & 5{,}666 (912) & & & & & \\
\bottomrule
\end{tabular}}
\end{table}

\begin{figure}[!htbp]
\centering
\includegraphics[width=0.92\linewidth]{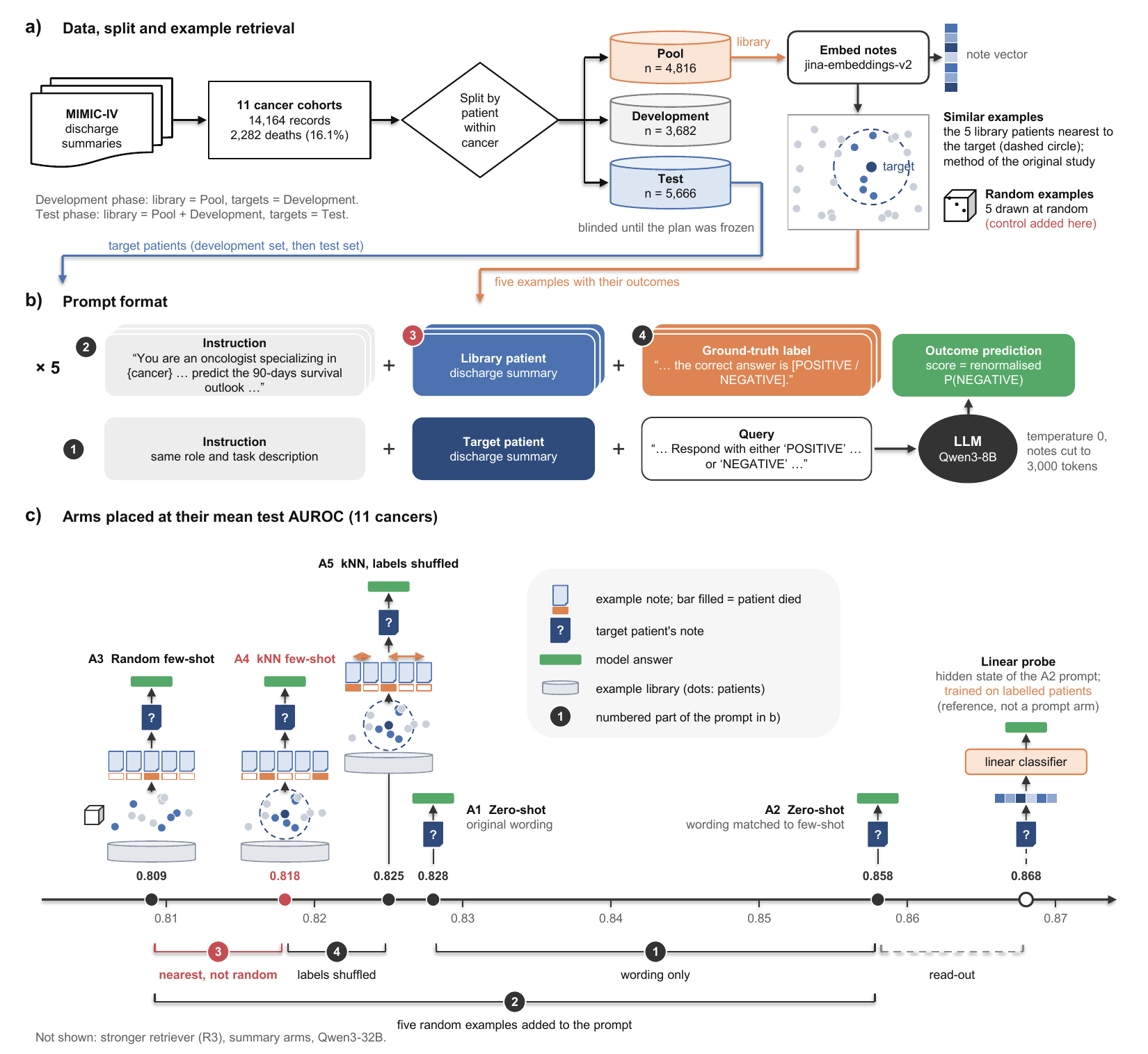}
\caption{\textbf{Study design and prompting strategies.} (a) Each of 11 MIMIC-IV cancer cohorts was split by patient into an example pool, a development set and a held-out test set; test-set metrics for the language model were first evaluated after the analysis plan had been frozen. In-context examples were either the five library patients nearest to the target in embedding space (the original method) or five patients drawn at random (the control added here). (b) Format of the few-shot prompt: five example blocks, each comprising an instruction, a library patient's discharge summary and the true outcome, followed by the target patient's block. The risk score is the renormalised probability assigned to NEGATIVE, the answer that indicates death. The numbers mark the four parts of the prompt that the strategies vary. (c) The five main prompting strategies and the hidden-state probe, placed at their mean test-set AUROC over the 11 cancers. Brackets connect strategies that differ in a single numbered part of the prompt: wording (1), presence of examples (2), choice of examples (3; the primary comparison) and the outcomes attached to the examples (4). The probe reads the hidden state of the zero-shot prompt with matched wording (A2) using a classifier trained on labelled patients; it is a reference, not a prompting strategy. In c, ``arms'' are strategies and ``kNN'' marks similar examples. The axis shows unweighted means without uncertainty; paired differences with confidence intervals are given in Table~\ref{tab:contrasts} and Fig.~\ref{fig:forest}. Dots and outcome patterns are schematic.}
\label{fig:design}
\end{figure}

We evaluated dynamic few-shot prompting with Qwen3-8B in a paired design in which every strategy was applied to the same patients (Fig.~\ref{fig:design}). Two zero-shot prompts presented the target patient's note alone, one with the zero-shot wording of the original implementation (A1) and one with wording matched to the few-shot prompt (A2). Few-shot prompts placed five labelled patients from the example library before the target note: the five nearest neighbours of the target in the embedding space of the original study (A4; dynamic few-shot prompting), five patients drawn at random (A3) or, as a further control, the five nearest neighbours with their outcomes permuted among them (A5). The risk score was the renormalised probability that the model assigned to the answer indicating death, and discrimination, the ranking of patients who died above survivors, was summarised by the AUROC. Differences between strategies were estimated within each cancer and pooled by inverse-variance weighting, whereas the mean AUROC of a strategy is the unweighted average over cancers and is descriptive. Intervals are 95\% confidence intervals.

\subsection*{Similarity did not improve risk ranking}

\begin{figure}[!htbp]
\centering
\includegraphics[width=\linewidth]{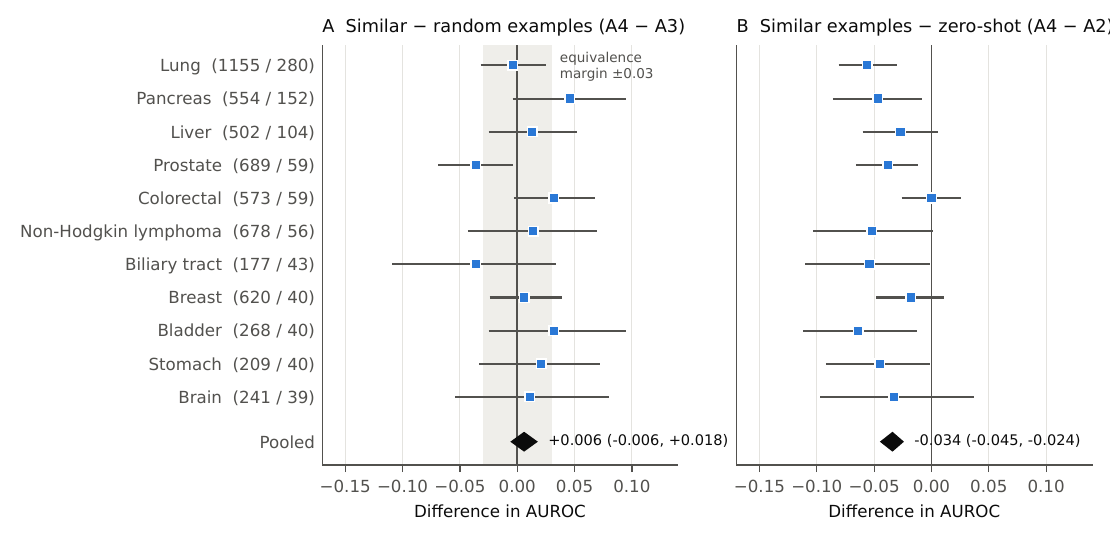}
\caption{\textbf{Per-cancer and pooled differences in AUROC on the held-out test set.} (A) Similar minus random examples (A4 $-$ A3; the primary contrast); the shaded band marks the pre-specified equivalence margin of $\pm0.03$. (B) Similar examples minus the patient's note alone with matched wording (A4 $-$ A2). Squares are cancer-specific paired differences with 95\% joint-bootstrap intervals, and diamonds are inverse-variance pooled estimates. Numbers in parentheses are patients / deaths.}
\label{fig:forest}
\end{figure}

The primary analysis asked whether examples selected for their similarity to the target ranked patients by risk of death better than examples selected at random. With the retrieval encoder of the original method, they did not. The pooled difference in AUROC between similar and random examples was $+0.006$ ($-0.006$ to $0.018$; Table~\ref{tab:contrasts}, Fig.~\ref{fig:forest}A). The whole interval lay within the pre-specified equivalence margin of $\pm0.03$, establishing equivalence under our decision rule, and also included zero, so no difference was detected. Mean AUROC over the 11 cancers was 0.818 with similar and 0.809 with random examples (Extended Data Table~\ref{tab:auroc}). Similar examples gave the numerically higher AUROC in 8 of 11 cancers, but the cancer-specific interval excluded zero only for prostate cancer, in which similar examples performed worse ($-0.036$; $-0.069$ to $-0.004$). Heterogeneity across cancers was modest ($I^2=28$\%), and the difference between similar and random examples was unrelated to the number of deaths in each test cohort (Spearman correlation 0.03). In this setting, selecting examples by semantic similarity contributed little to risk discrimination beyond what arbitrary labelled examples provided.

\begin{table}[!htbp]
\centering
\caption{\textbf{Pooled paired contrasts on the held-out test set (Qwen3-8B, full notes).} Differences in AUROC (accuracy for the threshold contrast) with 95\% joint-bootstrap intervals. ``Zero-shot'' denotes the prompt with matched wording unless stated otherwise, and ``similar examples'' those retrieved with the original encoder unless stated otherwise. ``Cancers $>0$'' is the number of cancers with a positive difference and $I^2$ the heterogeneity across cancers. Holm-adjusted $p$ values are given for the secondary family only.}
\label{tab:contrasts}
\small
\resizebox{\linewidth}{!}{%
\begin{tabular}{lrcccc}
\toprule
Contrast & Difference & 95\% CI & Cancers $>0$ & $I^2$ (\%) & Adjusted $p$ \\
\midrule
\multicolumn{6}{l}{\emph{Primary analysis}} \\
Similar $-$ random examples & $+0.006$ & $-0.006$, $0.018$ & 8/11 & 28 & -- \\
\midrule
\multicolumn{6}{l}{\emph{Secondary analyses (Holm-adjusted over 12 contrasts)}} \\
Random examples $-$ zero-shot & $-0.043$ & $-0.054$, $-0.032$ & 0/11 & 59 & $<0.001$ \\
Similar examples $-$ zero-shot & $-0.034$ & $-0.045$, $-0.024$ & 1/11 & 30 & $<0.001$ \\
Similar examples, true $-$ shuffled outcomes & $-0.006$ & $-0.015$, $0.003$ & 3/11 & 0 & $0.50$ \\
Zero-shot with fitted threshold $-$ similar examples (accuracy) & $+0.024$ & $0.016$, $0.031$ & 11/11 & 54 & $<0.001$ \\
Zero-shot $-$ TF-IDF + LR & $+0.014$ & $0.003$, $0.025$ & 8/11 & 61 & $0.059$ \\
Similar examples $-$ TF-IDF + LR & $-0.020$ & $-0.032$, $-0.007$ & 6/11 & 58 & $0.013$ \\
Zero-shot $-$ Qwen3-Embedding + LR & $-0.006$ & $-0.016$, $0.005$ & 5/11 & 61 & $0.50$ \\
Similar examples $-$ Qwen3-Embedding + LR & $-0.041$ & $-0.053$, $-0.028$ & 3/11 & 65 & $<0.001$ \\
Pooled probe $-$ zero-shot & $+0.011$ & $0.004$, $0.017$ & 8/11 & 42 & $0.004$ \\
Pooled probe $-$ similar examples & $+0.045$ & $0.034$, $0.056$ & 11/11 & 29 & $<0.001$ \\
Pooled probe $-$ Qwen3-Embedding + LR & $+0.006$ & $-0.004$, $0.015$ & 6/11 & 42 & $0.50$ \\
Pooled probe $-$ TF-IDF + LR & $+0.025$ & $0.015$, $0.035$ & 10/11 & 49 & $<0.001$ \\
\midrule
\multicolumn{6}{l}{\emph{Exploratory analyses (not adjusted)}} \\
Zero-shot, original $-$ matched wording & $-0.025$ & $-0.033$, $-0.018$ & 1/11 & 80 & -- \\
Similar examples, Qwen3-Embedding $-$ original retriever & $+0.017$ & $0.006$, $0.029$ & 7/11 & 36 & -- \\
Similar examples (Qwen3-Embedding) $-$ random examples & $+0.021$ & $0.010$, $0.033$ & 9/11 & 32 & -- \\
Similar examples (Qwen3-Embedding) $-$ zero-shot & $-0.019$ & $-0.028$, $-0.009$ & 1/11 & 0 & -- \\
Qwen3-Embedding + LR $-$ TF-IDF + LR & $+0.022$ & $0.012$, $0.032$ & 10/11 & 40 & -- \\
Qwen3-Embedding + LR $-$ jina + LR & $+0.038$ & $0.029$, $0.047$ & 9/11 & 47 & -- \\
\bottomrule
\end{tabular}}
\end{table}

\EDtable
\begin{table}[!htbp]
\centering
\caption{\textbf{Risk discrimination on the held-out test set.} Mean AUROC is the unweighted mean over the 11 cancers and is descriptive. Inference rests on the paired difference from the reference (ref.) of each block, the zero-shot prompt with matched wording under the same model and input, pooled over cancers by inverse variance; supervised comparators are compared with the Qwen3-8B full-note reference. ``Labelled records'' is the number of labelled patients on which a method draws: five per prediction for few-shot prompts, the cancer-specific library (264 to 1{,}733 records) for the classifiers and 8{,}126 records across cohorts for the pooled probe. Qwen3-32B results are based on 5{,}665 records. ``--'', not estimated. LR, logistic regression.}
\label{tab:auroc}
\small
\resizebox{\linewidth}{!}{%
\begin{tabular}{llcccc}
\toprule
Method & Labelled records & Mean AUROC & $\dauc$ vs.\ ref. & 95\% CI & Cancers above ref. \\
\midrule
\multicolumn{6}{l}{\emph{Qwen3-8B, full notes}} \\
Zero-shot, original wording (A1) & 0 & 0.828 & $-0.025$ & $-0.033$, $-0.018$ & 1/11 \\
Zero-shot, matched wording (A2) & 0 & 0.858 & ref. &  & -- \\
Random examples (A3) & 5 & 0.809 & $-0.043$ & $-0.054$, $-0.032$ & 0/11 \\
Similar examples, original retriever (A4) & 5 & 0.818 & $-0.034$ & $-0.045$, $-0.024$ & 1/11 \\
Similar examples, outcomes shuffled (A5) & 5 (permuted) & 0.825 & -- & -- & -- \\
Similar examples, Qwen3-Embedding retriever (R3) & 5 & 0.836 & $-0.019$ & $-0.028$, $-0.009$ & 1/11 \\
\midrule
\multicolumn{6}{l}{\emph{Qwen3-8B, summarised notes}} \\
Zero-shot, summary alone (S0) & 0 & 0.854 & ref. &  & -- \\
Random summarised examples (S3) & 5 & 0.839 & -- & -- & -- \\
Similar summarised examples (S4) & 5 & 0.847 & $-0.007$ & $-0.014$, $0.000$ & 3/11 \\
\midrule
\multicolumn{6}{l}{\emph{Qwen3-32B, full notes}} \\
Zero-shot, matched wording & 0 & 0.871 & ref. &  & -- \\
Random examples & 5 & 0.855 & $-0.013$ & $-0.020$, $-0.005$ & 1/11 \\
Similar examples, original retriever & 5 & 0.848 & $-0.022$ & $-0.030$, $-0.013$ & 1/11 \\
\midrule
\multicolumn{6}{l}{\emph{Supervised comparators}} \\
TF-IDF + LR & library & 0.831 & $-0.014$ & $-0.025$, $-0.003$ & 3/11 \\
jina + LR & library & 0.815 & -- & -- & -- \\
Qwen3-Embedding + LR & library & 0.855 & $+0.006$ & $-0.005$, $0.016$ & 6/11 \\
Pooled hidden-state probe & 8{,}126 & 0.868 & $+0.011$ & $0.004$, $0.017$ & 8/11 \\
\bottomrule
\end{tabular}}
\end{table}

\subsection*{Examples lowered discrimination}

\begin{figure}[!htbp]
\centering
\includegraphics[width=\linewidth]{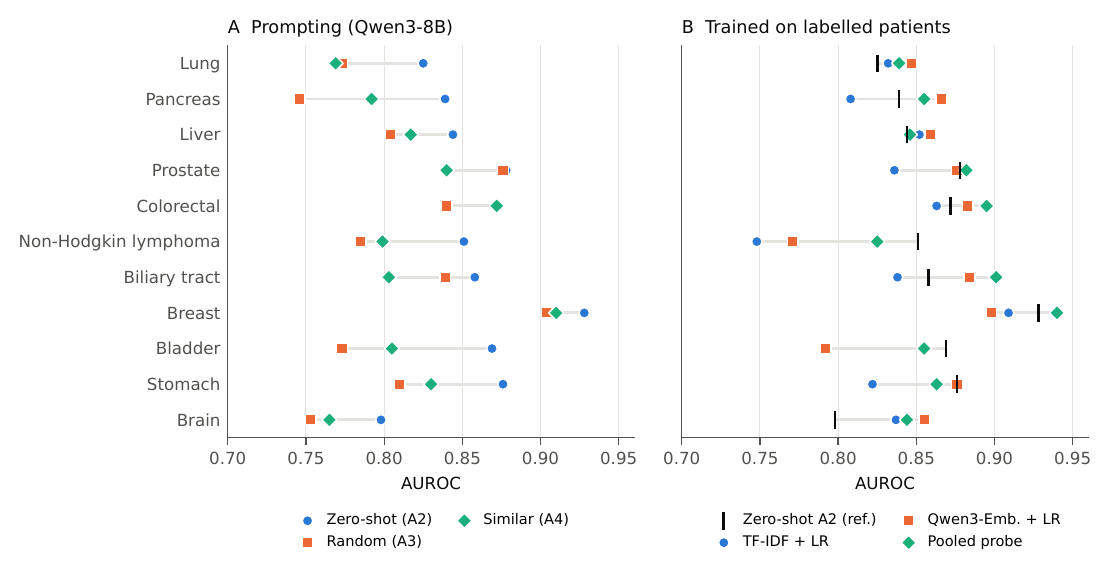}
\caption{\textbf{Test-set AUROC by cancer for prompting strategies and supervised comparators.} (A) Prompting with Qwen3-8B: the patient's note alone with matched wording (A2), random examples (A3) and similar examples (A4). (B) Classifiers trained on labelled records: TF-IDF + logistic regression (LR), Qwen3-Embedding + LR and the pooled hidden-state probe; the black tick marks the zero-shot prompt (A2) for reference. The methods in (B) use labelled training data and are shown as reference points, not as like-for-like alternatives to prompting.}
\label{fig:dots}
\end{figure}

We next asked whether labelled examples of either kind improved on the patient's note alone. To isolate the effect of the examples from that of prompt wording, the reference for this comparison was the zero-shot prompt with matched wording. Both few-shot strategies lowered AUROC. The pooled difference from the note alone was $-0.034$ ($-0.045$ to $-0.024$) for similar examples and $-0.043$ ($-0.054$ to $-0.032$) for random examples (Holm-adjusted $p<0.001$ for both; Table~\ref{tab:contrasts}). Random examples yielded a lower AUROC than the note alone in every cancer, and similar examples in ten; in the remaining cancer (colorectal) the difference was 0.000 (Figs.~\ref{fig:forest}B and~\ref{fig:dots}A).

The wording of the question affected discrimination even when no examples were shown. Mean AUROC was 0.828 with the original zero-shot wording and 0.858 with the matched wording, a pooled difference of $-0.025$ ($-0.033$ to $-0.018$; exploratory; $I^2=80$\%). Comparing a few-shot prompt with the original zero-shot prompt, as in the original design, therefore sets it against the weaker of the two zero-shot prompts in our data. Even so, both few-shot strategies had a lower mean AUROC than the original zero-shot prompt, a comparison that was not among the planned contrasts.

Permuting the outcome labels among the five retrieved patients did not reduce discrimination (true minus shuffled labels, $-0.006$; $-0.015$ to $0.003$; adjusted $p=0.50$). This contrast is diluted by construction, because for 2{,}738 of the 5{,}666 test records (48\%) all five neighbours shared one outcome, almost always survival, and shuffling left the prompt unchanged. Within that limit, the result provides no evidence that the model exploited the link between an example's note and its outcome.

\subsection*{Retriever, summaries and model size}

Because a null result for similarity could reflect a weak retriever rather than the absence of a benefit, we repeated dynamic prompting with Qwen3-Embedding-8B, an encoder whose embeddings carried more prognostic information than those of the original encoder (Table~\ref{tab:contrasts}). Examples retrieved with this encoder (R3) achieved a higher AUROC than examples retrieved with the original encoder ($+0.017$; $0.006$ to $0.029$) and than random examples ($+0.021$; $0.010$ to $0.033$; both exploratory). The latter interval excludes zero and extends beyond the equivalence margin, so the equivalence of similar and random examples is specific to the retriever of the original method. Discrimination nevertheless remained below that of the note alone ($-0.019$; $-0.028$ to $-0.009$): a stronger retriever recovered roughly half of the loss associated with examples, but not all of it.

The best-performing configuration of the original study combined retrieved examples with LLM-written summaries of the notes. In the closest analogue available to us, in which Qwen3-8B summarised every note under six fixed headings (Supplementary Note~\ref{supp:summary}), summarisation raised the AUROC of dynamic prompting by 0.028 ($0.018$ to $0.038$) with no detectable change in zero-shot discrimination ($-0.002$; $-0.010$ to $0.005$), and similar summarised examples remained marginally below the summary alone ($-0.007$; $-0.014$ to $0.000$) and within the $\pm0.03$ margin of random summarised examples without a detectable difference ($+0.007$; $-0.002$ to $0.016$).

Model scale changed the size of these effects but not their direction. With AWQ-quantised Qwen3-32B, the note alone reached a mean AUROC of 0.871, examples again reduced discrimination, by 0.022 ($0.013$ to $0.030$) for similar and by 0.013 ($0.005$ to $0.020$) for random examples, and similar examples were not better than random ones (Supplementary Note~\ref{supp:scale}). In no configuration that we examined did examples improve on the patient's note alone.

\subsection*{Supervised classifiers and a probe}

To place the prompting results on a common scale, we compared them with text classifiers trained on the labelled library of each cancer, which comprised 264 to 1{,}733 records (Extended Data Table~\ref{tab:auroc}, Fig.~\ref{fig:dots}B). The zero-shot prompt performed comparably with these classifiers although it used no labels, with no statistically significant difference detected after adjustment relative to logistic regression on TF-IDF features ($+0.014$; $0.003$ to $0.025$; adjusted $p=0.059$) or on Qwen3-Embedding vectors ($-0.006$; $-0.016$ to $0.005$; adjusted $p=0.50$). Dynamic prompting, in contrast, was inferior to both ($-0.020$ and $-0.041$; adjusted $p=0.013$ and $p<0.001$; Table~\ref{tab:contrasts}). On the held-out test set, a linear probe trained on the final-layer hidden state of the zero-shot prompt and pooled across cancers (8{,}126 training records) had a mean AUROC of 0.868 and exceeded the model's own zero-shot answer in paired comparisons by 0.011 ($0.004$ to $0.017$; adjusted $p=0.004$). Supervised linear decoding of the same representation thus yielded only a small gain in risk discrimination over the zero-shot answer.

\subsection*{Accuracy rose as death predictions fell}

Classification at a fixed cut-off gave a different picture. When death was predicted for scores of at least 0.5, adding examples raised accuracy although it lowered AUROC (Table~\ref{tab:accuracy}). Unweighted mean accuracy over the 11 cancers was 46.9\% with the original zero-shot wording, 75.1\% with the matched wording, 82.7\% with random examples and 82.8\% with similar examples. The accuracy gain reported for dynamic few-shot prompting was therefore reproduced in our data, but it was the same for random as for similar examples. It was accompanied by a marked change in how often the model predicted death: the mean proportion of patients predicted to die fell from 69\% with the original zero-shot wording to 36\% with the matched wording and to 18\% with either kind of example, against an observed mortality of 16\%. Consistent with this shift, examples reduced the Brier score from 0.227 for the note alone to 0.144 with similar and 0.145 with random examples; because the Brier score reflects resolution as well as calibration, this reduction does not show that the scores are calibrated risks.

\begin{table}[!htbp]
\centering
\caption{\textbf{Accuracy on the held-out test set (\%).} Accuracy at a threshold of 0.5 for the two zero-shot prompts and the two few-shot strategies, and for the zero-shot prompt with matched wording after applying one threshold per cancer fitted to development-set scores. ``Majority'' predicts survival for every patient. The highest value in each row is in bold. The last row gives the mean proportion of patients predicted to die.}
\label{tab:accuracy}
\small
\resizebox{\linewidth}{!}{%
\begin{tabular}{lcccccc}
\toprule
 & & \multicolumn{2}{c}{Zero-shot} & \multicolumn{2}{c}{Few-shot} & Zero-shot (A2) \\
\cmidrule(lr){3-4}\cmidrule(lr){5-6}
Cancer type & Majority & Original (A1) & Matched (A2) & Random (A3) & Similar (A4) & + threshold \\
\midrule
Lung & 75.8 & 40.8 & 64.5 & 75.9 & 74.6 & \textbf{80.0} \\
Pancreas & 72.6 & 29.1 & 39.2 & 67.3 & 72.6 & \textbf{78.0} \\
Liver & 79.3 & 30.7 & 71.3 & 80.7 & 80.5 & \textbf{84.3} \\
Prostate & 91.4 & 61.4 & 88.1 & 91.3 & 90.6 & \textbf{92.0} \\
Colorectal & 89.7 & 58.5 & 85.9 & 87.8 & 89.7 & \textbf{91.4} \\
Non-Hodgkin lymphoma & 91.7 & 51.9 & 89.7 & 92.3 & 91.9 & \textbf{93.2} \\
Biliary tract & 75.7 & 29.4 & 65.0 & 77.4 & 76.8 & \textbf{80.2} \\
Breast & \textbf{93.5} & 72.6 & 87.7 & 92.7 & 91.8 & 93.1 \\
Bladder & 85.1 & 60.1 & 86.2 & 84.7 & 85.1 & \textbf{88.4} \\
Stomach & 80.9 & 36.4 & 72.2 & 78.5 & 78.5 & \textbf{84.7} \\
Brain (malignant) & 83.8 & 45.2 & 75.9 & 81.3 & 78.4 & \textbf{86.3} \\
\midrule
Mean over cancers & 83.6 & 46.9 & 75.1 & 82.7 & 82.8 & 86.5 \\
Predicted to die, mean (\%) & -- & 68.8 & 35.7 & 18.4 & 18.1 & 8.9 \\
\bottomrule
\end{tabular}}
\end{table}

If the accuracy gain reflects a shift in decision tendency rather than better ranking, moving the threshold should reproduce it without examples. It did. A single threshold per cancer, fitted to zero-shot scores in the development set and applied unchanged to the test set, raised mean accuracy to 86.5\% and exceeded the accuracy of both few-shot strategies in all 11 cancers (pooled difference from similar examples $+2.4$ percentage points; $1.6$ to $3.1$; adjusted $p<0.001$). Fitting these thresholds, like showing examples, requires labelled patients. A label-free alternative examined post hoc, a variant of batch calibration\citep{zhou2024batch} that predicted death when a patient's score exceeded the mean score of the test patients with the same cancer, did not reproduce the gain: accuracy was 74.7\%, lower than with similar examples in all 11 cancers, and the rule still predicted death for 36\% of patients.

Higher accuracy did not, however, signify better identification of the patients who died. Predicting survival for every patient gave a mean accuracy of 83.6\%, which similar examples exceeded in only 3 of 11 cancers and random examples in 4. The accuracy-maximising thresholds were themselves extreme (0.998 or higher on the score scale) and predicted death for only 9\% of patients; with them, balanced accuracy was 0.65, against 0.75 for the zero-shot prompt at 0.5 and 0.69 for similar examples (Supplementary Table~\ref{tab:metrics_test}). At these event rates accuracy rewards predicting few deaths, and neither its increase with examples nor its further increase with a fitted threshold identifies a better decision rule.

\subsection*{Answers track the outcomes displayed}

\begin{figure}[!htbp]
\centering
\includegraphics[width=\linewidth]{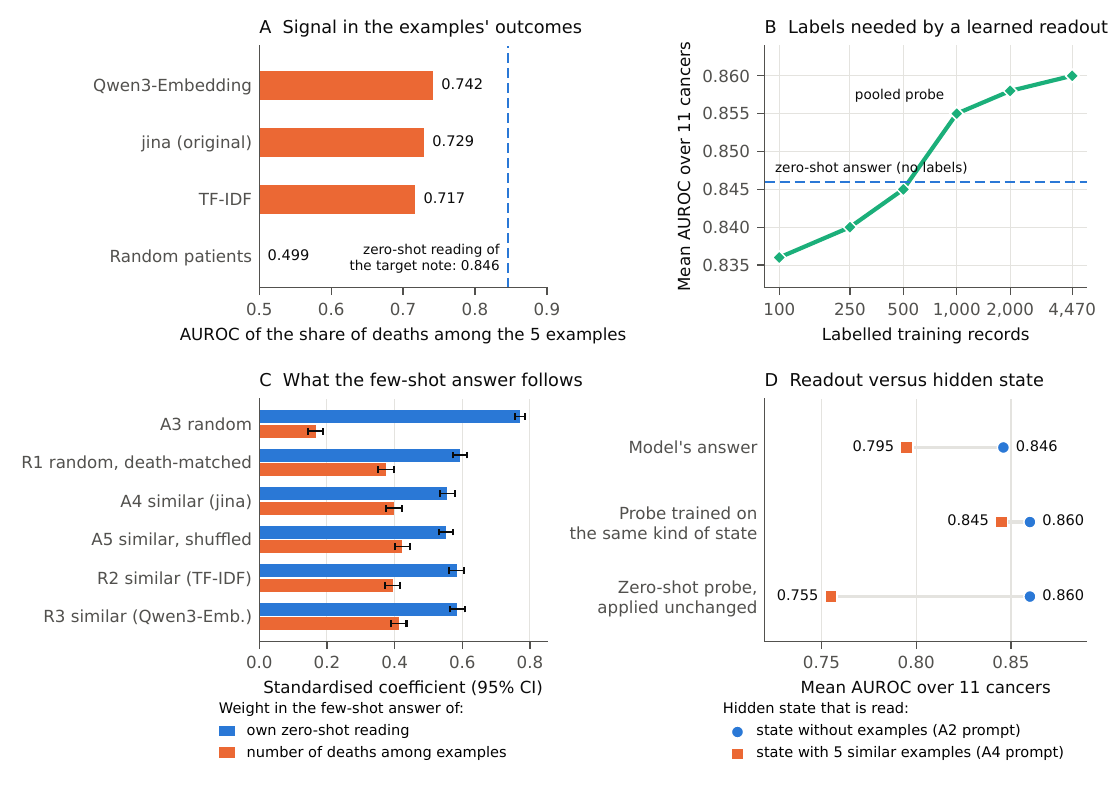}
\caption{\textbf{Retrieved examples carry prognostic signal but shift the model's answer towards the outcomes displayed (development set).} (A) AUROC of the neighbour vote, the proportion of deaths among the five selected library patients, for random selection and three retrievers (unweighted mean over 11 cancers); the dashed line marks the zero-shot prompt (A2). (B) Mean AUROC of the pooled hidden-state probe against the number of labelled training records; the dashed line marks the zero-shot prompt, which uses no labels. (C) Standardised coefficients of Eq.~\eqref{eq:anchor} with 95\% intervals for the six few-shot strategies: weight of the model's own zero-shot reading (blue) and of the number of deaths among the examples (orange). (D) Mean AUROC of three readouts applied to the hidden state without examples (blue circles) and with five similar examples (orange squares). Throughout, blue refers to the model reading the target note without examples and orange to quantities that involve the examples.}
\label{fig:mechanism}
\end{figure}

To understand why similar examples did not help, we first verified that retrieval identified patients who resembled the target (Extended Data Table~\ref{tab:neighbours_test}). Compared with random library patients, the five neighbours retrieved by the original encoder more often shared the target's admission type (43\% versus 24\%) and discharge destination (39\% versus 27\%), although the encoder had not received these variables as separate fields. Their outcomes were also prognostic. Used as a risk score, the proportion of deaths among a target's five neighbours, a neighbour vote that involves no language model, gave a mean AUROC of 0.72 for the original encoder and 0.75 for Qwen3-Embedding, against 0.52 for random patients, with the same ordering in the development set (Fig.~\ref{fig:mechanism}A). Retrieval therefore worked as intended, but the neighbours' outcomes were considerably less informative than the target's own note, from which the model reached an AUROC of 0.86 without any examples.

\EDtable
\begin{table}[!htbp]
\centering
\caption{\textbf{Properties of the five library patients selected for each target on the test set.} Values are unweighted means over the 11 cancers, by selection method; the library is the pool plus the development set. ``Neighbour-vote AUROC'' uses the proportion of deaths among the five selected patients as the risk score. ``Outcome agreement'' is the proportion of selected patients with the same 90-day outcome as the target. The last two rows count targets, out of 5{,}666, whose five selected patients share one outcome.}
\label{tab:neighbours_test}
\small
\begin{tabular}{lcccc}
\toprule
 & Random & TF-IDF & jina (original) & Qwen3-Embedding \\
\midrule
Neighbour-vote AUROC & 0.515 & 0.712 & 0.722 & 0.752 \\
Outcome agreement & 0.739 & 0.785 & 0.767 & 0.793 \\
Same admission type & 0.235 & 0.471 & 0.434 & 0.432 \\
Same discharge location & 0.272 & 0.415 & 0.387 & 0.412 \\
Same sex & 0.613 & 0.702 & 0.690 & 0.728 \\
Absolute age difference (years) & 14.6 & 12.8 & 12.6 & 12.1 \\
\midrule
Targets with five survivors & 2{,}603 & 3{,}069 & 2{,}692 & 3{,}145 \\
Targets with five deaths & 3 & 18 & 46 & 38 \\
\bottomrule
\end{tabular}
\end{table}

We then examined how the examples entered the model's answer. Within each cancer, we regressed the few-shot log-odds of death on the zero-shot log-odds for the same patient and on the number of deaths among the five examples shown, with all variables standardised (Methods). If examples left the model's reading of the target note intact, the coefficient of the zero-shot log-odds would be 1 and that of the death count 0. Instead, with similar examples the weight of the model's own reading fell to 0.55 ($0.53$ to $0.56$) on the test set, while the death count received a weight of 0.43 ($0.41$ to $0.45$), and shuffling the outcome labels left both coefficients essentially unchanged (Supplementary Table~\ref{tab:anchor}). With random examples, the death count still carried a positive weight of 0.17 ($0.16$ to $0.19$). Because random examples were drawn independently of the target, the number of deaths among them varied by chance, so this coefficient indicates that showing more patients who died moved the answer towards death irrespective of the target patient. Development-set estimates were nearly identical (Fig.~\ref{fig:mechanism}C), and in a further control run in that set, random patients matched to the neighbours on the number of deaths displayed reproduced the coefficients of dynamic prompting and achieved a higher AUROC than dynamic prompting in all 11 cancers ($+0.019$; $0.003$ to $0.034$).

Together, these observations suggest an account of the main results: in-context examples add the displayed death count as an input to the answer and reduce the weight given to the target note. With random examples the added input is noise. With similar examples it is a genuine signal, the neighbour vote, but a weaker one (AUROC of about 0.72) than the reading it partly displaces (0.86). Either substitution would lower discrimination relative to the note alone and bring the two few-shot strategies close to each other. This account was formed after inspection of the development data, and except for the random-example coefficient the regression describes associations, not the effects of interventions.

Hidden states from the development set help to locate the loss (Fig.~\ref{fig:mechanism}B,D; Supplementary Note~\ref{supp:mech}). Without examples, the model's zero-shot answer behaved as a fixed linear readout of its final-layer state (mean correlation 0.988 with the projection onto the answer direction), and a probe fitted to recorded outcomes read the same state only slightly better (mean AUROC 0.860 against 0.846). When five similar examples preceded the note, the mean AUROC of the model's answer fell to 0.795, whereas a probe trained on the corresponding few-shot states reached 0.845. Examples thus lowered the AUROC of the model's answer far more than that of a linear readout fitted to the hidden state, a pattern consistent with a change in readout direction rather than a loss of linearly decodable signal, although the few-shot states also encode the examples and their outcomes.

\subsection*{Exclusion of hospice discharges}

Because discharge to hospice is stated in the summary and strongly predicts death, we repeated the contrasts after excluding the 176 test records (3.1\%) with this disposition, 149 of which were followed by death within 90 days. Mean AUROC decreased for every method, as expected when the most predictable patients are removed (from 0.858 to 0.840 for the zero-shot prompt with matched wording), but the conclusions were unchanged (Supplementary Table~\ref{tab:nohospice_test}): similar and random examples remained equivalent ($+0.005$; $-0.009$ to $0.019$), both remained below the note alone ($-0.039$; $-0.051$ to $-0.027$; and $-0.047$; $-0.060$ to $-0.034$), and the probe retained its advantage over the zero-shot answer ($+0.010$; $0.003$ to $0.017$).

% =====================================================================
\section*{Discussion}

In this controlled re-examination of dynamic few-shot prompting across 11 cancer cohorts, the three components that the method changes at once had separable and partly opposing effects. Selecting examples by similarity, the component that motivates the approach, did not improve risk discrimination over random selection when the retrieval encoder of the original method was used. The presence of labelled examples of either kind reduced discrimination relative to the patient's note alone, and the wording of the prompt altered both discrimination and, more strongly, the proportion of patients predicted to die. At the same time, examples raised accuracy at the default threshold, which reproduces the principal benefit reported for the method. The two observations are reconciled by distinguishing how well a prompt ranks patients from how readily it predicts death: examples corrected a pronounced tendency of the zero-shot prompt to over-predict death, and a threshold fitted to zero-shot scores achieved a larger accuracy gain without them. These conclusions were consistent between the development and test sets, were unchanged after the exclusion of patients discharged to hospice and held with a larger model.

The dissociation between accuracy and discrimination aligns with what is known about in-context learning in classification more generally: a prompt carries an implicit prior over answers, and both its wording and the labels of its demonstrations shift that prior\citep{zhao2021calibrate}. Because the zero-shot prompt with matched wording predicted death more than twice as often as it occurred (36\% versus 16\%), curbing that over-prediction substantially reduced false-positive calls and raised accuracy at the 0.5 threshold even as risk ranking worsened. This correction is not without value, since examples brought the predicted death rate close to the observed rate and lowered the Brier score from 0.23 to 0.14, but a fitted threshold provided it more effectively, and at the event rates studied here it did not translate into better identification of the patients who died. The original study noted that full-note zero-shot predictions for glioma fell almost entirely into one class, a setting in which accuracy is particularly sensitive to a model's tendency to favour one answer.

Why did similar patients not help the model to rank risk? Retrieval did not fail: the retrieved neighbours resembled the target, and their outcomes alone predicted the target's outcome with an AUROC above 0.7. The target's own discharge summary was, however, a considerably stronger source of prognostic information, and the regression analysis suggests that, when examples are present, the model's answer follows the number of deaths displayed at the expense of its own reading of the target note. A model that shifts towards the outcomes of its neighbours can be expected to rank patients less well when its reading of the target is the better signal, and may gain when the two signals are comparable. The same reasoning accounts for the proximity of similar and random examples, which displace part of a strong signal by a weaker one or by noise, and for the partial recovery with the stronger retriever. Because shuffling preserves the number of deaths displayed, its null result, in line with reports that the correctness of in-context labels matters less than expected\citep{min2022rethinking}, does not test anchoring to that number. Prompt length is an alternative explanation for the loss: five example notes lengthened the prompt from about 2{,}700 to about 16{,}000 tokens, and models use long inputs less reliably\citep{liu2024lost}. The smaller loss with summaries is compatible with both explanations, and a zero-shot prompt lengthened with outcome-free text, which we did not run, would help to separate them.

Our findings should not be read as contradicting the original report\citep{anil2026dynamic}, whose cohorts, endpoints and models differed from ours and whose data we could not re-examine (Extended Data Table~\ref{tab:reconcile}; Supplementary Note~\ref{supp:original}). Its accuracy gain has a direct counterpart in our data, and its gains in discrimination were themselves mixed: with the 8B model they were not significant, whereas with the 70B model the AUROC for glioma increased from 0.742 to 0.813. Several differences could explain why ranking improved there but worsened in our cohorts: the language model; the balance of outcomes, since for 48\% of our test records the five neighbours shared one outcome; the relative informativeness of the target note, which in our cohorts far exceeded that of the neighbour vote; the outcome horizon, which in the original study lay months to years after diagnosis; and the form of the examples, which in the best original configuration were summaries written to one template per cancer. Retrieved examples may therefore help most when the zero-shot model strongly favours one answer, when the outcome is defined relative to the cohort, when the examples are short and uniformly structured, and when outcomes are balanced, so that the neighbours are about as informative as the target note. A random-example control in the original cohorts would determine whether similarity contributed to the gains reported there.

\EDtable
\begin{table}[!htbp]
\centering
\caption{\textbf{Findings of the original study and their counterparts in our test set.} Figures for the original study are taken from its tables.}
\label{tab:reconcile}
\small
\begin{tabular}{p{5.2cm}p{6.6cm}p{2.6cm}}
\toprule
Original study & This study (test set) & Agreement \\
\midrule
Accuracy rises with dynamic examples (glioma, 8B, five summarised examples versus full-note zero-shot: 65.7\% to 75.4\%) & Accuracy rises from 75.1\% (zero-shot, matched wording) to 82.8\% with similar examples & Yes \\
Summaries of target and example notes, with a different number of examples, outperform full notes for glioma but not for breast cancer (70B) & Summaries of target and example notes give a 0.028 higher AUROC with similar examples & Mixed \\
AUROC with five summarised examples versus full-note zero-shot. 8B: breast 0.743 to 0.791 and glioma 0.748 to 0.775, both non-significant. 70B: breast 0.765 to 0.741, non-significant, and glioma 0.742 to 0.813 ($p<0.05$ in fold-level tests that the authors treat as descriptive) & AUROC falls by 0.034 with full notes and by 0.007 with summaries & Mixed \\
Similar examples are better than no examples; no random-example control & Similar and random examples are equivalent with the original retriever; similar examples are 0.021 better with a stronger one & Not tested in the original \\
\bottomrule
\end{tabular}
\end{table}

These results have implications for the design of studies that evaluate similar-patient retrieval. Four checks would make such studies easier to interpret: a random-example control with the same number of examples and the same prompt; a zero-shot baseline whose wording matches the few-shot prompt; a threshold-free metric reported together with accuracy, the proportion of positive predictions and the accuracy of always predicting the majority outcome; and a zero-shot baseline with a threshold fitted on separate labelled data. The first three require no additional patients. Zero-shot prompting was also a strong baseline in its own right: without labelled data and with an input about one sixth as long as the full-note few-shot prompts, it ranked patients about as well as logistic regressions trained on several hundred to more than a thousand labelled records per cancer. Its raw scores should nevertheless not be used as probabilities or thresholded at 0.5, and the accuracy-maximising thresholds are not a clinical alternative, as they raised accuracy while lowering balanced accuracy from 0.75 to 0.65. Any use of such scores, for instance to prioritise patients for earlier follow-up or for discussions of supportive care after discharge, would require calibration on local data, with the operating point chosen for the intended clinical purpose and not for accuracy.

Several limitations merit discussion. All cohorts came from a single centre, the input was a single discharge summary and the outcome was 90-day mortality; summaries that state prognosis and disposition outright may leave little room for examples to help, and the findings may not transfer to less informative notes, to longer horizons or to the longitudinal notes and long-term outcomes of the original study. We studied one model family, and the comparison between Qwen3-8B and AWQ-quantised Qwen3-32B does not isolate model size from quantisation. Several design decisions were made after development data had been seen: the equivalence margin was set after results for three cancers were available, all development results were known before the analysis plan was frozen, and the stronger retriever was added to the test run after its development result had been seen; an early TF-IDF baseline for lung cancer was also computed on the full cohort, although no language-model output for test patients was examined. The split was random rather than temporal, and the development and test sets come from the same cohorts, so agreement between them is not external validation. Notes were cut to 3{,}000 tokens, our summaries were written by the 8B model without clinician review, each strategy was run once at temperature 0 with one random draw of examples per patient, cancer types were defined by diagnosis codes in any position, and no clinician reviewed the cohort definitions, the prompts, the summaries or the retrieved cases. We did not compare prompting with models built on structured variables such as age, admission type and discharge destination, and we varied neither the number of examples nor prompt length independently of the examples. The analyses of neighbours, anchoring and hidden states were designed after the development results were known; whereas the pooled zero-shot hidden-state probe was tested on the held-out test set, the supplementary analyses of training-label number and of hidden states with and without examples used the development set only, and the death-matched and TF-IDF-retrieved few-shot strategies were not run on the test set. Finally, calibration was examined only through accuracy-maximising thresholds and, post hoc, a batch-calibration variant, so neither the raw answer probabilities nor the renormalised scores should be read as calibrated risks; the pooled estimates are fixed-effect combinations with considerable heterogeneity for some contrasts ($I^2$ up to 80\%), and the retriever, wording, summary and fusion results are exploratory.

Overall, in 11 cancer cohorts from MIMIC-IV, showing an LLM the notes and outcomes of similar patients did not improve its ranking of patients by risk of death within 90 days of discharge. Similar examples selected with the retriever of the original method were equivalent within the pre-specified margin to random examples, both kinds of example worsened risk ranking relative to the patient's note alone, and their accuracy gain at the default threshold came from predicting fewer deaths, a gain that a cancer-specific threshold on note-alone scores exceeded. Future evaluations of retrieval-based prompting for clinical prediction should include a random-example control and a wording-matched zero-shot baseline, so that the contribution of similarity can be separated from those of wording and of adding examples.

% =====================================================================
\section*{Methods}

\subsection*{Study design and relation to the original study}

This study was designed as a conceptual replication of the dynamic prompting framework of \citet{anil2026dynamic}, extended with controls that isolate its components. Because the original cohorts are not public, we retained the retrieval encoder, the prompt wording and the procedure by which the model's answer is converted into a score, and changed the data, the outcome, the language model and the summariser (Extended Data Table~\ref{tab:vs_original}). Absolute performance therefore cannot be compared between the two studies, and our results do not bear on the accuracy of the figures reported for the original cohorts. Within our cohorts, similar examples were compared with random examples to test the contribution of similarity, prompts with examples were compared with the patient's note alone under matched wording to test the contribution of adding examples, and true outcomes were compared with shuffled outcomes for the same similar examples to test the contribution of the link between each example and its outcome.

\EDtable
\begin{table}[!htbp]
\centering
\caption{\textbf{Design of the original study and of this study.}}
\label{tab:vs_original}
\small
\begin{tabular}{p{3.3cm}p{5.6cm}p{6.1cm}}
\toprule
 & \citet{anil2026dynamic} & This study \\
\midrule
Data & One centre, not public; breast cancer ($n=503$) and glioma ($n=475$) & MIMIC-IV, credentialed access under a data use agreement; 11 cancer types, 14{,}164 records \\
Input & All notes from the first 180 days after diagnosis & One discharge summary \\
Outcome & Survival beyond 5 years (breast) or 14 months (glioma) from diagnosis & All-cause death within 90 days after the prediction time, the same for all cancers \\
Evaluation & Stratified five-fold cross-validation & Fixed pool, development and test split; language-model test metrics evaluated after the analysis plan was frozen \\
Language model & Llama-3 (3.0) 8B and 70B Instruct, 262k-context variants & Qwen3-8B; Qwen3-32B as a control for model size \\
Summaries & GPT-4 Turbo; about one third sampled for clinician review, discrepancies corrected if found & Qwen3-8B, no human review \\
Examples & $k=2$ with full notes, $k=5$ with summaries & $k=5$ with full notes and with summaries \\
Zero-shot wording & Differs from the few-shot wording in the public code\citep{anil2025dynamiccode} & Both the original wording and a wording matched to the few-shot prompt \\
Random-example control & No & Yes \\
Shuffled-outcome control & No & Yes \\
\bottomrule
\end{tabular}
\end{table}

\subsection*{Data source and cohort selection}

We used MIMIC-IV version 3.1\citep{johnson2023mimiciv,johnson2024mimiciv31} together with the linked free-text notes in MIMIC-IV-Note version 2.2\citep{johnson2023mimicivnote}, which contain de-identified records of patients admitted to a single academic medical centre in Boston. For each cancer type, we assembled a cohort in which each patient contributed one index admission, defined as the earliest admission that satisfied all of the following criteria: (1) the patient was at least 18 years old at admission; (2) the patient did not die in hospital; (3) the admission carried an ICD-9 or ICD-10 diagnosis code for the cancer in any diagnosis position (Supplementary Table~\ref{tab:icd}); and (4) the admission had a non-empty discharge summary with valid chart and storage timestamps. Starting from a list of 25 cancer groups defined by ICD code prefixes, we retained every group with at least 100 deaths within 90 days, which yielded ten groups, and added malignant brain tumours (98 deaths) as the closest available counterpart to the glioma cohort of the original study, giving 11 cancer types (Extended Data Table~\ref{tab:cohort}). Because a qualifying code could occupy any diagnosis position, the cancer was not necessarily the reason for admission. The brain tumour group is defined by codes for malignant neoplasm of the brain and is not restricted to glioma, and cohort sizes reflect admissions to one hospital rather than population incidence.

A patient with codes for more than one cancer contributed one record to each corresponding cohort, possibly from different index admissions, and we therefore count patient--cancer records. The cohorts contained 14{,}164 records and 2{,}282 deaths, with deaths also counted by record. The records belonged to 13{,}351 unique patients, of whom 779 appeared in more than one cohort (746 in two, 32 in three and one in four).

\subsection*{Prediction time and outcome definition}

Each prediction was anchored at the time at which the patient had left hospital and the summary supplied to the model was available. Because discharge and summary availability can occur at different times, the prediction time $t_0$ was defined as the later of the two, a summary being considered available at the later of its chart and storage times. If an admission had more than one eligible summary, the one that became available first was used. Patients whose recorded date of death fell on or before the date of $t_0$ were excluded. The outcome was all-cause death on calendar days 1--90 after the date of $t_0$, and death is the positive class in all analyses. Patients with a recorded death after day 90 were classified as survivors. For patients without a recorded death, follow-up was taken to extend to one calendar year after their last recorded hospital discharge in MIMIC-IV; they were classified as survivors if this date reached day 90 and were excluded otherwise, which affected very few patients (for example, one in the lung cohort).

\subsection*{Data partitioning}

Each cohort was divided once, by patient and with stratification by outcome, into an example pool, a development set and a test set. The test set comprised 40\% of each cohort. For ten cancers the remaining patients were divided equally between the development set and the pool (30\% each); for lung cancer the development set comprised 300 patients, including 200 who had been used in a pilot run, and the pool received the rest. The pool supplied the examples placed in prompts and the training data for all supervised comparators, the development set was used to debug the pipeline, to choose among design options and for exploratory analyses, and the test set was reserved for the final evaluation.

During development, examples and training data came from the pool only. For the test phase, the library of labelled cases available as examples or as training data was the union of the pool and the development set. This was decided before any test inference, because some pools were small (181 patients for brain tumours and 465 for breast cancer), which could have handicapped similarity-based retrieval; as the libraries differ, absolute performance is not compared between the two phases. Cohorts were split separately, so a patient who appeared in two cohorts could be a training case for one cancer and an evaluation case for another, and analyses that combine records across cancers were designed to account for this.

\subsection*{Input notes}

The only input for a patient was the text of the discharge summary of the index admission. Summaries longer than 3{,}000 tokens of the Qwen3-8B tokenizer were shortened by retaining the first 2{,}000 and the last 1{,}000 tokens, joined by a marker (\texttt{[...]}); the final portion was retained because the end of a discharge summary usually contains the hospital course, the discharge diagnoses and the discharge disposition. About half of all notes were shortened in this way (37--66\% by cancer; Extended Data Table~\ref{tab:cohort}), and the same shortened text was used whether a note belonged to the target patient or served as an example.

Discharge summaries also state where the patient was discharged to, and discharge to hospice was strongly associated with the outcome: in the full lung cohort, 103 of 121 patients discharged to hospice died within 90 days, and the proportion was between 85\% and 91\% in nine of the 11 cohorts (69\% for breast cancer and 50\% for brain tumours). These patients, who made up 1--6\% of each cohort, were retained in the main analysis because the discharge destination is part of the summary available at $t_0$, and were removed in a sensitivity analysis.

\subsection*{Language models and risk score}

The main language model was Qwen3-8B\citep{yang2025qwen3}, run with vLLM\citep{kwon2023vllm} at temperature 0 and with the model's thinking mode disabled. The Llama-3 models of the original study were not available to us. To assess dependence on model scale, three strategies were repeated with AWQ-quantised Qwen3-32B (Supplementary Note~\ref{supp:scale}). The same note also describes the inference hardware and compares the scores obtained on the two computing set-ups.

Following the original implementation, the model was asked to answer \textsc{positive} if the patient was likely to survive beyond 90 days and \textsc{negative} otherwise; the answer \textsc{negative} therefore indicates death, although death is the positive class in all our statistics. The model's reply was pre-filled with ``\texttt{The correct answer is }'', generation was limited to four tokens, and the 20 most probable tokens at each position were recorded. The risk score was the probability of \textsc{negative} as a proportion of the probability of the two answers together, so that a higher score indicates death more strongly:
\begin{equation}
\pneg \;=\; \frac{P_{\textsc{neg}}}{P_{\textsc{neg}} + P_{\textsc{pos}}},
\label{eq:score}
\end{equation}
where $P_{\textsc{neg}}$ and $P_{\textsc{pos}}$ sum the probabilities of all returned tokens that, after removal of whitespace, quotation marks and asterisks, begin with \texttt{NEG} or \texttt{POS}. This renormalisation does not make the score a calibrated risk of death. The original implementation scores the renormalised probability of \textsc{positive} against survival\citep{anil2025dynamiccode}; our score is its complement, scored against death, which leaves AUROC unchanged. The score was read at the first generated position at which the most probable token was one of the two answers. This was never the first position, because after the pre-filled prefix Qwen3-8B always emitted an opening quotation mark or Markdown emphasis before the answer word. A patient was predicted to die if $\pneg \ge 0.5$. Every Qwen3-8B response could be scored, in all strategies and on both sets; one Qwen3-32B response could not, and the corresponding record was excluded from the Qwen3-32B analyses.

\subsection*{Prompting strategies}

All strategies listed in Extended Data Table~\ref{tab:arms} for a given set were run on the same patients, so that every comparison is paired. Zero-shot strategies show the model the target patient's note alone, and few-shot strategies place labelled examples before it. Fig.~\ref{fig:design} shows the data flow, the prompt format and the part of the prompt in which each strategy differs from its comparator, and the full prompt texts are given in Supplementary Note~\ref{supp:prompts}.

\EDtable
\begin{table}[!htbp]
\centering
\caption{\textbf{Prompting strategies.} All strategies used Qwen3-8B. ``Dev'' and ``Test'' indicate the sets on which a strategy was run. A2--A4 were repeated on the test set with Qwen3-32B.}
\label{tab:arms}
\small
\begin{tabular}{llp{7.6cm}cc}
\toprule
ID & Strategy & In-context content & Dev & Test \\
\midrule
A1 & Zero-shot, original wording & None; zero-shot wording of the original code & \checkmark & \checkmark \\
A2 & Zero-shot, matched wording & None; few-shot target prompt with the references to examples removed & \checkmark & \checkmark \\
A3 & Random examples & Five library patients drawn uniformly at random, with true outcomes & \checkmark & \checkmark \\
A4 & Similar examples & Five nearest library patients by jina embedding (the original method), with true outcomes & \checkmark & \checkmark \\
A5 & Similar, outcomes shuffled & Same five patients as A4; their outcomes permuted among them & \checkmark & \checkmark \\
R1 & Random, death-matched & Five random library patients with as many deaths as the A4 examples of the same target & \checkmark & -- \\
R2 & Similar by TF-IDF & Five nearest library patients by TF-IDF cosine similarity & \checkmark & -- \\
R3 & Similar by Qwen3-Embedding & Five nearest library patients by Qwen3-Embedding-8B & \checkmark & \checkmark \\
S0 & Summary, zero-shot & None; as A2 with the note replaced by its summary & \checkmark & \checkmark \\
S3 & Summary, random examples & As A3 with every note replaced by its summary & \checkmark & \checkmark \\
S4 & Summary, similar examples & As A4 with every note replaced by its summary & \checkmark & \checkmark \\
\bottomrule
\end{tabular}
\end{table}

The few-shot prompt followed that of the original code sentence by sentence; only the cancer name, the horizon (``90 day'') and the time point (``at hospital discharge'') were changed. Each example consisted of the task preamble, the example patient's note and the line ``For this example, the correct answer is \textsc{positive}/\textsc{negative}''; the five examples were followed by the task description, the target note and the question, and similar examples were listed in decreasing order of similarity. Five examples ($k=5$) were used throughout, whereas the original study used $k=5$ with summaries and, because of context length, $k=2$ with full notes. On the development set, a zero-shot prompt contained about 2{,}700 tokens on average and a five-shot prompt about 16{,}000.

Two zero-shot prompts separated the effect of examples from that of wording. The zero-shot prompt in the original code is shorter than the few-shot prompt, describes the role differently and does not explain what \textsc{positive} and \textsc{negative} mean, so that comparing it with a few-shot prompt changes the wording and the presence of examples at once. The first zero-shot prompt (A1) reproduced this original wording, and the second (A2) was the few-shot target prompt from which only the two phrases that refer to examples were removed. The effect of examples was measured against A2, and the effect of wording alone as the difference between A1 and A2.

Three controls varied the examples. In the first, the similar patients (A4) were replaced by patients drawn uniformly at random, without replacement, from the same library (A3), with the random generator seeded per target patient; similar minus random examples (A4 $-$ A3) is the primary contrast. In the second, the five similar patients and their texts were retained but their outcome labels were permuted among them (A5), which keeps the number of deaths shown and breaks only the link between each note and its outcome; when all five similar patients have the same outcome, the prompts with true and shuffled outcomes are identical. The third addressed the fact that the similar patients of a high-risk target include more deaths than a random draw, so that similar and random examples also differ in the mix of outcomes shown. In a control run on the development set only (R1), random patients were drawn such that, for each target, the number of deaths among the five examples matched that of the similar examples; the example patients were not matched individually.

\subsection*{Retrieval of similar patients}

As in the original study, each patient was represented by an embedding of the note computed with \texttt{jina-embeddings-v2-base-en}\citep{gunther2023jina}. The full, unshortened note was embedded, and notes exceeding the encoder's 8{,}192-token window were split into chunks whose normalised embeddings were averaged. The similar examples of a target were the five library patients with the highest cosine similarity to it, which we also refer to as its neighbours.

Because a null result for similarity could reflect a weak retriever, the similar-example strategy was repeated with two other retrievers. The first (R2) selected neighbours by the cosine similarity of TF-IDF vectors (unigrams and bigrams, sublinear term frequency, at most 50{,}000 features, fitted on the library). The second (R3) used Qwen3-Embedding-8B\citep{zhang2025qwen3embedding} on up to 8{,}192 tokens per note, without an instruction prefix; neighbours were ranked by cosine similarity and no reranker was applied. On the development set, a logistic regression on Qwen3-Embedding vectors predicted the outcome better than one on jina vectors ($\dauc$ $+0.041$; $0.028$ to $0.054$; Supplementary Note~\ref{supp:dev}), so that by this measure R3 retrieves with a representation that carries more prognostic information. R3 was added to the test run as a robustness analysis after this development result had been seen.

\subsection*{Note summarisation}

The best configuration of the original study combined five similar examples with LLM-written summaries of the notes, and we evaluated the closest analogue available to us. Qwen3-8B summarised each shortened note under six headings (diagnosis and stage; treatment history; hospital course and complications; symptoms and functional status; molecular or laboratory markers; and discharge disposition and follow-up plan), using only information stated in the note, in no more than 250 words (generation limit 420 tokens). Three strategies repeated the full-note comparisons with every note replaced by its summary: the target summary alone (S0), with five random examples (S3) or with five similar examples (S4). Similar patients were still retrieved with the jina embedding of the full note, so that only the text shown to the model changed; in the original study, retrieval in the summary configuration appears to have used embeddings of the summaries themselves. Unlike in the original study, the same model wrote the summaries and made the predictions, no clinician reviewed the summaries, and about 6\% of them reached the token limit and were cut off (Supplementary Note~\ref{supp:summary}).

\subsection*{Supervised comparators}

Supervised text classifiers were trained to place the prompting results on a common scale. They used the same library that supplied the in-context examples and were trained separately for each cancer, on 264 to 1{,}733 labelled records in the test phase. The first was a logistic regression with balanced class weights on TF-IDF features (regularisation parameter $C=1$, not tuned). The second was a logistic regression on standardised jina or Qwen3-Embedding-8B vectors, with balanced class weights and $C$ chosen from $\{10^{-3},10^{-2},10^{-1}\}$ by five-fold cross-validation within the training data. A majority rule that predicts survival for every patient served as the reference for accuracy; as it assigns the same prediction to all patients, it cannot discriminate between them. An exploratory fusion of the zero-shot score with the TF-IDF score is described in Supplementary Note~\ref{supp:fusion}.

\subsection*{Hidden-state probe}

To assess how well outcome risk can be linearly decoded from the model's internal representation when it reads a note without examples, we trained a linear probe\citep{alain2017understanding} on its hidden states. The rationale is that the model's zero-shot answer is itself, to a close approximation, a fixed linear readout of the hidden state. Let $h_i \in \mathbb{R}^d$ be the final-layer hidden state of the last prompt token for patient $i$ under the zero-shot prompt with matched wording. A language model obtains its next-token logits by multiplying this state with its output embedding matrix, so that the log-odds of answering \textsc{negative} rather than \textsc{positive} are, up to the handling of spelling variants and of the position at which the answer appears, a linear function of $h_i$ along a direction that is fixed by pre-training:
\begin{equation}
z_i^{\mathrm{zs}} \;=\; \log\frac{P_{\textsc{neg}}}{P_{\textsc{pos}}} \;\approx\; u^{\top} h_i ,
\qquad u = w_{\textsc{neg}} - w_{\textsc{pos}},
\label{eq:readout}
\end{equation}
where $w_{\textsc{neg}}$ and $w_{\textsc{pos}}$ are the output-embedding rows of the two answer tokens. Zero-shot prompting thus scores a patient with a direction that was never fitted to the outcome. The probe keeps the same input and replaces $u$ by a direction fitted to recorded outcomes:
\begin{equation}
\hat p_i \;=\; \sigma\!\left(a^{\top} \tilde h_i^{(\ell)} + b\right),
\qquad
(a,b) \;=\; \arg\min_{a,b}\; \sum_{i \in \mathcal{T}} c_{y_i}\,
\log\!\left(1 + e^{-(2y_i-1)\,(a^{\top}\tilde h_i^{(\ell)} + b)}\right) + \frac{1}{2C}\lVert a\rVert_2^2 ,
\label{eq:probe}
\end{equation}
where $\tilde h_i^{(\ell)}$ is the standardised hidden state at layer $\ell$, $y_i\in\{0,1\}$ indicates death, $\mathcal{T}$ is the training set and $c_{y}$ are class weights inversely proportional to class frequency. This is an $\ell_2$-regularised logistic regression, and the language model is not changed. When $\ell$ is the final layer, Eqs.~\eqref{eq:readout} and~\eqref{eq:probe} read the same state, up to standardisation, with a direction that is given ($u$) or learned ($a$), and the difference in AUROC measures the gain from the fitted readout.

States were extracted at layers $\ell \in \{12,\allowbreak 18,\allowbreak 24,\allowbreak 28,\allowbreak 32,\allowbreak 36\}$ of 36, and the layer and $C\in\{10^{-3},\allowbreak 10^{-2},\allowbreak 10^{-1}\}$ were chosen by AUROC on out-of-fold predictions from five-fold cross-validation within the training set (Supplementary Note~\ref{supp:mech}). Because per-cancer probes were unstable in small cohorts, a single probe was pooled across cancers. During development it was trained on the pools of all 11 cancers, excluding every patient who was in the development or test set of any cancer (4{,}470 records, including 757 deaths); cross-validation selected the final layer and $C=10^{-3}$. For the test phase, the layer and $C$ were fixed at these values and the probe was trained on the pools and development sets of all cancers, excluding every patient in the test set of any cancer (8{,}126 records, including 1{,}303 deaths).

The probe is fitted on several thousand labelled records, whereas the zero-shot prompt uses no labels and the few-shot prompts see five labelled patients per query. It is therefore an empirical reference for linear decoding from the model's state after reading the note, and it was used as a diagnostic for the prompting results and not as a proposed predictor.

\subsection*{Analyses of retrieved neighbours and of the influence of examples}

Three exploratory analyses examined whether the retrieved patients resembled the target and how the examples changed the model's answer. First, for each target we compared the five retrieved patients with five random library patients with respect to their outcomes, to structured variables that the retriever had not received as separate fields (admission type, discharge location, sex and age) and to cosine similarity in the two embedding spaces. As a direct check of prognostic value, the proportion of deaths among a target's five selected patients was used as that target's risk score and its AUROC was computed; this neighbour vote uses only the selected patients' outcomes and is not produced by the language model.

Second, we measured how closely each few-shot answer followed the model's answer from the patient's note alone and the number of deaths $d_i$ among the five examples shown. Both answers were expressed as log-odds of \textsc{negative} against \textsc{positive}, and within each cancer the few-shot log-odds were regressed on the zero-shot log-odds and $d_i$, with all variables standardised:
\begin{equation}
z_i^{\mathrm{fs}} \;=\; \beta_0 + \beta_{\mathrm{zs}}\, z_i^{\mathrm{zs}} + \beta_{d}\, d_i + \varepsilon_i .
\label{eq:anchor}
\end{equation}
Here $\beta_{\mathrm{zs}}$ is the weight of the model's own reading of the note and $\beta_d$ the weight of the death count. Coefficients were pooled over the 11 cancers by inverse variance, with variances from 1{,}000 bootstrap resamples. In every strategy except random examples, the death count comes from the target's neighbours and is correlated with the target's risk, so that part of $\beta_d$ may reflect legitimate use of that information; with random examples the count is independent of the target by construction, but its different distribution means that the size of its standardised coefficient cannot be compared with that of the other strategies (Supplementary Note~\ref{supp:mech}).

Third, we extracted the hidden state of the last prompt token under the similar-example prompt for pool and development patients and trained the pooled probe on these few-shot states with the same training records and regularisation as before. Zero-shot and few-shot states were also projected onto the answer direction $u$ of Eq.~\eqref{eq:readout}, and the zero-shot probe was repeated after removal of the component of each state along $u$.

These analyses were designed after the development results were known and while the test set was still blinded. As specified in the frozen analysis plan, the neighbour analysis and the regression were repeated on the test set, whereas these supplementary hidden-state comparisons (and the training-label curve in Supplementary Note~\ref{supp:mech}) used the development set only.

\subsection*{Evaluation metrics and statistical analysis}

AUROC was the main metric. It measures ranking, namely the probability that a randomly chosen patient who died has a higher score than a randomly chosen survivor, with ties counted as one half, and it does not depend on a decision threshold. AUROC, the area under the precision--recall curve (AUPRC) and the Brier score use continuous scores, whereas accuracy, balanced accuracy, the F1 score for death and the proportion of patients predicted to die depend on classification at a decision threshold, which was 0.5 unless stated otherwise.

Within each cancer, the difference between two strategies was estimated on the same patients, with a confidence interval from a paired bootstrap over patients (2{,}000 resamples; percentile interval). The 11 cancer-specific differences were combined as an inverse-variance weighted mean with variances taken from the bootstrap (a fixed-effect combination). The mean AUROC reported for a strategy, in contrast, gives all 11 cancers equal weight, so the pooled difference between two strategies can differ from the difference between their mean AUROCs. On the test set, patients who belonged to more than one cohort (152 of 5{,}514 unique patients) were resampled jointly across cohorts, and the pooled interval is the percentile interval of the pooled estimate over the bootstrap replicates, which reflects the dependence between cohorts. The development-set analyses resampled each cancer independently and used a normal approximation for the pooled interval, which may make those intervals slightly too narrow. Alongside each pooled estimate we report the number of cancers in which the difference is positive and the heterogeneity statistic $I^2$, the estimated proportion of the variation in the cancer-specific differences that is due to heterogeneity rather than sampling error.

Two strategies were considered different if the 95\% interval of the pooled difference excluded zero, and equivalent if the interval lay entirely within $\pm0.03$ AUROC; equivalence was judged on the whole interval because an interval that includes zero can also arise from an imprecise comparison that leaves larger differences plausible. Both criteria can be met by the same interval, and if neither is met the result is reported as inconclusive; for contrasts not evaluated against this equivalence margin, an interval including zero (or an adjusted $p\ge0.05$) indicates failure to detect a difference rather than equivalence. The margin was fixed before any test inference, when development results for the first three cancers had been seen; from the width of the development-set intervals we expected a half-width of about 0.02 for the pooled primary contrast on the test set.

The comparison of similar with random examples was the single primary test. Twelve secondary contrasts (Table~\ref{tab:contrasts}) addressed the effect of examples relative to the note alone, the role of the examples' true outcomes, the substitution of examples by a decision threshold, the comparison of prompting with supervised classifiers and the comparison with the hidden-state probe; they were adjusted with Holm's method\citep{holm1979simple}, using $p$ values from a normal approximation based on the bootstrap standard deviation of the pooled difference. The three contrasts with Qwen3-32B formed a separate family, and a descriptive analysis related the primary difference to cohort size. All other contrasts are exploratory and are reported without adjustment.

To test whether a change of decision threshold could replace the accuracy gain from examples, we fitted a single accuracy-maximising threshold to the scores from the patient's note alone with matched wording. If the best finite threshold was strictly less accurate, on the data used to fit it, than predicting survival for every patient, that prediction was used instead; if the two tied, the finite threshold was kept. On the development set the threshold was cross-fitted in five folds, and for the test set it was fitted per cancer on development-set scores and applied unchanged. This procedure moves the cut-off and does not turn the scores into calibrated risks. A variant of batch calibration\citep{zhou2024batch}, which predicts death when a patient's score exceeds the mean score of the test patients with the same cancer, was examined post hoc as a label-free alternative; batch calibration as originally described subtracts the batch mean in log-probability space, which for two classes corresponds to a threshold at the mean log-odds rather than at the mean score. All contrasts were also repeated after the removal of patients discharged to hospice.

\subsection*{Pre-specification and blinding}

The primary comparison, the equivalence margin and the main prompting strategies (A1--A5) were chosen after development results for three cancers had been seen and before any inference on the test set. Development results for all 11 cancers were known by the time the analysis plan, comprising the list of contrasts, the multiplicity family and the statistical methods, was frozen. Test-set inference stored one score per patient and strategy and computed no quantity involving the outcome labels, and the stored language-model scores were first compared with the true outcomes after the plan had been frozen. The Qwen3-Embedding retriever (R3) was added on development evidence while blinded test inference was running. The Qwen3-32B runs were completed after the Qwen3-8B results were known; their scores were kept blinded until the runs were complete and were analysed once with the frozen code. Analyses added after unblinding are labelled post hoc.

\subsection*{Ethical approval and data governance}

MIMIC-IV and MIMIC-IV-Note are de-identified resources that are available to credentialed users through PhysioNet under a data use agreement\citep{goldberger2000physionet}. The collection of patient information and the creation of the research resource were reviewed by the Institutional Review Board at the Beth Israel Deaconess Medical Center, which granted a waiver of informed consent and approved the data-sharing initiative\citep{johnson2023mimiciv,johnson2024mimiciv31}. The present analysis was conducted under the PhysioNet credentialed data use agreement. In line with PhysioNet's guidance on the use of MIMIC data with LLMs\citep{physionet2025llm}, all processing and all model inference were performed on local hardware or on a rented GPU server under our control, and no patient text was sent to any external API or third-party service.

\subsection*{Data availability}
MIMIC-IV (version 3.1; \url{https://doi.org/10.13026/kpb9-mt58}) and MIMIC-IV-Note (version 2.2; \url{https://doi.org/10.13026/1n74-ne17}) are available to credentialed users through PhysioNet. No patient-level data are released with this study.

\subsection*{Code availability}
The analysis code, the cohort definitions and the scripts that regenerate the patient-level splits from the source files are held in a private GitHub repository. Requests for access may be directed to the authors.

\bibliographystyle{unsrtnat}
\bibliography{refs}

\section*{Acknowledgements}

\section*{Author contributions}

\section*{Competing interests}

\clearpage
\section*{Extended Data}
\printEDtables

% =====================================================================
% Supplementary Information
% =====================================================================
\clearpage
\setcounter{table}{0}
\setcounter{figure}{0}
\renewcommand{\tablename}{Supplementary Table}
\renewcommand{\figurename}{Supplementary Fig.}
\renewcommand{\theHtable}{supp.\arabic{table}}
\renewcommand{\theHfigure}{supp.\arabic{figure}}

\begin{center}
{\Large\bfseries Supplementary Information}
\end{center}

% ---------------------------------------------------------------------
\suppnote{Cohort definitions and prompt templates}{supp:prompts}

Supplementary Table~\ref{tab:icd} lists the ICD code prefixes that define each cancer type. Malignant brain tumour codes are not equivalent to glioma, the liver group mixes histologies, a diagnosis code does not imply that the cancer was the reason for admission, and cohort size does not reflect incidence.

\begin{table}[!htbp]
\centering
\caption{\textbf{ICD code prefixes defining each cancer type.} An admission qualifies if it carries a listed code in any diagnosis position.}
\label{tab:icd}
\small
\begin{tabular}{lll}
\toprule
Cancer type & ICD-9 prefixes & ICD-10 prefixes \\
\midrule
Lung                 & 162.2--162.5, 162.8, 162.9 & C34 \\
Colorectal           & 153, 154.0, 154.1 & C18, C19, C20 \\
Pancreas             & 157 & C25 \\
Stomach              & 151 & C16 \\
Liver and intrahepatic bile duct & 155 & C22 \\
Biliary tract (gallbladder and other) & 156 & C23, C24 \\
Bladder              & 188 & C67 \\
Prostate             & 185 & C61 \\
Breast               & 174, 175 & C50 \\
Brain (malignant)    & 191 & C71 \\
Non-Hodgkin lymphoma & 200, 202.0--202.2, 202.7, 202.8 & C82--C86 \\
\bottomrule
\end{tabular}
\end{table}

The prompts below are the texts given to the model. Text in braces is filled in for each patient; \texttt{\{cancer\}} is the English name of the cancer type (for example, ``lung cancer'' or ``non-Hodgkin lymphoma''). Every prompt is wrapped in the Qwen3 chat template as a single user message with thinking disabled, and the assistant turn is pre-filled with ``\texttt{The correct answer is }''. In scoring, \textsc{negative} is the answer that indicates death and \textsc{positive} the answer that indicates survival beyond 90 days. Line breaks inside paragraphs are for typesetting only, and the grammar of the original prompts is kept as is.

\paragraph{Zero-shot prompt with the original wording (A1).}
{\small
\begin{verbatim}
You are an oncologist at a major cancer hospital, tasked with predicting outcomes
for patients.
I am going to provide you with a clinical notes for a {cancer} patient at hospital
discharge. Here is the summary: {target note}
Based on your understanding of the clinical notes for a {cancer} patient at hospital
discharge, classify the 90 day survival outlook for this patient.
Please respond with either POSITIVE or NEGATIVE
\end{verbatim}
}

\paragraph{Few-shot prompts (A3--A5, R1--R3).} The following block is repeated for each of the five examples:
{\small
\begin{verbatim}
You are an oncologist specializing in {cancer} at a major cancer hospital. Your task
is to predict the 90-days survival outlook for a {cancer} patient based on the
following clinical notes, which represents the patient's status at hospital discharge.

Here is the clinical notes example:
{example note}
For this example, the correct answer is {POSITIVE|NEGATIVE}.
\end{verbatim}
}
\noindent followed by the block for the target, the patient whose outcome is being predicted:
{\small
\begin{verbatim}
You are an oncologist specializing in {cancer} at a major cancer hospital. Your task
is to predict the 90 day survival outlook for a {cancer} patient based on the clinical
notes, which represents the patient's status at hospital discharge. I am going to
provide you previous patient examples to help guide the decision making.
Here are the clinical notes:
{target note}
Please analyze the clinical notes carefully, considering factors such as tumor
progression, treatment response, symptoms, and any relevant biomarkers. Based on this
analysis as well as the knowledge from the previous examples, classify the patient's
90 day survival outlook. Respond with either 'POSITIVE' (if the patient is likely to
survive beyond 90 days) or 'NEGATIVE' (if the patient is unlikely to survive beyond
90 days).
\end{verbatim}
}

\paragraph{Zero-shot prompt with matched wording (A2).} This prompt is the target block above without the sentence ``I am going to provide you previous patient examples to help guide the decision making.'' and without the phrase ``as well as the knowledge from the previous examples''. Nothing else is changed.

\paragraph{Summary-based strategies (S0, S3, S4).} These prompts are the same as A2, A3 and A4, respectively, with ``clinical notes'' replaced by ``clinical summary'' and each note replaced by its summary.

\paragraph{Summary generation.}
{\small
\begin{verbatim}
You are an oncologist. Summarize the following hospital discharge summary of a
{cancer} patient. Use only information stated in the note and do not guess. Use these
headings: Diagnosis and stage; Treatment history; Hospital course and complications;
Symptoms and functional status; Molecular or laboratory markers; Discharge disposition
and follow-up plan. Do not leave out clinical data relevant to prognosis. Keep it
under 250 words.

{note}
\end{verbatim}
}

% ---------------------------------------------------------------------
\clearpage
\suppnote{Additional test-set results}{supp:test}

Supplementary Tables~\ref{tab:percancer_prompt_test}--\ref{tab:percancer_contrasts_test} give the cancer-specific AUROCs and paired differences that underlie the pooled estimates in the main text. Supplementary Table~\ref{tab:metrics_test} reports additional metrics for all methods, and Supplementary Table~\ref{tab:nohospice_test} the sensitivity analysis without patients discharged to hospice.

\begin{table}[!htbp]
\centering
\caption{\textbf{Test-set AUROC by cancer for the prompting strategies (Qwen3-8B, full notes).}}
\label{tab:percancer_prompt_test}
\small
\begin{tabular}{lrcccccc}
\toprule
Cancer & $N$ / deaths & A1 & A2 & A3 & A4 & A5 & R3 \\
\midrule
Lung & 1155 / 280 & 0.777 & 0.825 & 0.773 & 0.769 & 0.768 & 0.806 \\
Pancreas & 554 / 152 & 0.732 & 0.839 & 0.746 & 0.792 & 0.812 & 0.806 \\
Liver & 502 / 104 & 0.776 & 0.844 & 0.804 & 0.817 & 0.831 & 0.813 \\
Prostate & 689 / 59 & 0.872 & 0.878 & 0.876 & 0.840 & 0.853 & 0.873 \\
Colorectal & 573 / 59 & 0.863 & 0.872 & 0.840 & 0.872 & 0.862 & 0.860 \\
Non-Hodgkin lymphoma & 678 / 56 & 0.868 & 0.851 & 0.785 & 0.799 & 0.803 & 0.844 \\
Biliary tract & 177 / 43 & 0.803 & 0.858 & 0.839 & 0.803 & 0.830 & 0.875 \\
Breast & 620 / 40 & 0.914 & 0.928 & 0.904 & 0.910 & 0.919 & 0.911 \\
Bladder & 268 / 40 & 0.834 & 0.869 & 0.773 & 0.805 & 0.835 & 0.846 \\
Stomach & 209 / 40 & 0.874 & 0.876 & 0.810 & 0.830 & 0.833 & 0.812 \\
Brain (malignant) & 241 / 39 & 0.796 & 0.798 & 0.753 & 0.765 & 0.731 & 0.749 \\
\midrule
Mean over cancers & & 0.828 & 0.858 & 0.809 & 0.818 & 0.825 & 0.836 \\
\bottomrule
\end{tabular}
\end{table}

\begin{table}[!htbp]
\centering
\caption{\textbf{Test-set AUROC by cancer for classifiers trained on labelled patients.} The zero-shot prompt with matched wording (A2) is shown for reference.}
\label{tab:percancer_sup_test}
\small
\resizebox{\linewidth}{!}{%
\begin{tabular}{lrccccc}
\toprule
Cancer & $N$ / deaths & A2 & TF-IDF + LR & jina + LR & Qwen3-Emb.\ + LR & Pooled probe \\
\midrule
Lung & 1155 / 280 & 0.825 & 0.832 & 0.798 & 0.847 & 0.839 \\
Pancreas & 554 / 152 & 0.839 & 0.808 & 0.814 & 0.866 & 0.855 \\
Liver & 502 / 104 & 0.844 & 0.852 & 0.814 & 0.859 & 0.846 \\
Prostate & 689 / 59 & 0.878 & 0.836 & 0.825 & 0.876 & 0.882 \\
Colorectal & 573 / 59 & 0.872 & 0.863 & 0.853 & 0.883 & 0.895 \\
Non-Hodgkin lymphoma & 678 / 56 & 0.851 & 0.748 & 0.721 & 0.771 & 0.825 \\
Biliary tract & 177 / 43 & 0.858 & 0.838 & 0.826 & 0.884 & 0.901 \\
Breast & 620 / 40 & 0.928 & 0.909 & 0.901 & 0.898 & 0.940 \\
Bladder & 268 / 40 & 0.869 & 0.791 & 0.802 & 0.792 & 0.855 \\
Stomach & 209 / 40 & 0.876 & 0.822 & 0.816 & 0.876 & 0.863 \\
Brain (malignant) & 241 / 39 & 0.798 & 0.837 & 0.792 & 0.855 & 0.844 \\
\midrule
Mean over cancers & & 0.858 & 0.831 & 0.815 & 0.855 & 0.868 \\
\bottomrule
\end{tabular}}
\end{table}

\begin{table}[!htbp]
\centering
\caption{\textbf{Cancer-specific differences in AUROC for the main contrasts on the test set,} with 95\% joint-bootstrap intervals.}
\label{tab:percancer_contrasts_test}
\small
\resizebox{\linewidth}{!}{%
\begin{tabular}{lrcrcrc}
\toprule
 & \multicolumn{2}{c}{A4 similar $-$ A3 random} & \multicolumn{2}{c}{A4 similar $-$ A2 zero-shot} & \multicolumn{2}{c}{A3 random $-$ A2 zero-shot} \\
\cmidrule(lr){2-3}\cmidrule(lr){4-5}\cmidrule(lr){6-7}
Cancer & $\dauc$ & 95\% CI & $\dauc$ & 95\% CI & $\dauc$ & 95\% CI \\
\midrule
Lung & $-0.004$ & $-0.032$, $0.025$ & $-0.056$ & $-0.081$, $-0.030$ & $-0.052$ & $-0.078$, $-0.025$ \\
Pancreas & $+0.046$ & $-0.004$, $0.095$ & $-0.047$ & $-0.086$, $-0.008$ & $-0.093$ & $-0.139$, $-0.048$ \\
Liver & $+0.013$ & $-0.025$, $0.052$ & $-0.027$ & $-0.060$, $0.006$ & $-0.041$ & $-0.074$, $-0.006$ \\
Prostate & $-0.036$ & $-0.069$, $-0.004$ & $-0.038$ & $-0.066$, $-0.012$ & $-0.002$ & $-0.027$, $0.026$ \\
Colorectal & $+0.032$ & $-0.003$, $0.068$ & $0.000$ & $-0.026$, $0.026$ & $-0.032$ & $-0.073$, $0.007$ \\
Non-Hodgkin lymphoma & $+0.014$ & $-0.043$, $0.070$ & $-0.052$ & $-0.103$, $0.001$ & $-0.065$ & $-0.114$, $-0.019$ \\
Biliary tract & $-0.036$ & $-0.109$, $0.034$ & $-0.054$ & $-0.110$, $-0.001$ & $-0.019$ & $-0.082$, $0.046$ \\
Breast & $+0.006$ & $-0.024$, $0.039$ & $-0.018$ & $-0.048$, $0.011$ & $-0.024$ & $-0.060$, $0.007$ \\
Bladder & $+0.032$ & $-0.025$, $0.095$ & $-0.064$ & $-0.112$, $-0.013$ & $-0.096$ & $-0.143$, $-0.057$ \\
Stomach & $+0.021$ & $-0.033$, $0.072$ & $-0.045$ & $-0.092$, $-0.001$ & $-0.066$ & $-0.115$, $-0.022$ \\
Brain (malignant) & $+0.011$ & $-0.054$, $0.080$ & $-0.033$ & $-0.097$, $0.037$ & $-0.044$ & $-0.099$, $0.012$ \\
\midrule
Pooled & $+0.006$ & $-0.006$, $0.018$ & $-0.034$ & $-0.045$, $-0.024$ & $-0.043$ & $-0.054$, $-0.032$ \\
\bottomrule
\end{tabular}}
\end{table}

\begin{table}[!htbp]
\centering
\caption{\textbf{Additional metrics on the test set (unweighted mean over the 11 cancers).} Accuracy, balanced accuracy, the F1 score for death and the proportion predicted to die use a threshold of 0.5, except for ``A2 + threshold''. The supervised classifiers use balanced class weights, which affects their threshold-dependent metrics.}
\label{tab:metrics_test}
\small
\resizebox{\linewidth}{!}{%
\begin{tabular}{lccccccc}
\toprule
Method & AUROC & AUPRC & Accuracy (\%) & Balanced acc. & F1 (death) & Brier & Predicted to die (\%) \\
\midrule
A1 Zero-shot, original wording & 0.828 & 0.451 & 46.9 & 0.660 & 0.362 & 0.504 & 68.8 \\
A2 Zero-shot, matched wording & 0.858 & 0.561 & 75.1 & 0.748 & 0.513 & 0.227 & 35.7 \\
A2 + threshold & -- & -- & 86.5 & 0.648 & 0.434 & -- & 8.9 \\
A3 Random examples & 0.809 & 0.477 & 82.7 & 0.693 & 0.483 & 0.145 & 18.4 \\
A4 Similar examples & 0.818 & 0.487 & 82.8 & 0.687 & 0.472 & 0.144 & 18.1 \\
A5 Similar, outcomes shuffled & 0.825 & 0.515 & 83.2 & 0.700 & 0.494 & 0.138 & 18.3 \\
R3 Similar (Qwen3-Embedding) & 0.836 & 0.530 & 83.5 & 0.700 & 0.500 & 0.136 & 18.0 \\
S0 Summary, zero-shot & 0.854 & 0.544 & 74.9 & 0.759 & 0.506 & 0.237 & 36.4 \\
S3 Summary, random & 0.839 & 0.517 & 80.6 & 0.747 & 0.508 & 0.173 & 25.4 \\
S4 Summary, similar & 0.847 & 0.541 & 82.6 & 0.747 & 0.530 & 0.158 & 23.2 \\
\midrule
TF-IDF + LR & 0.831 & 0.495 & 82.3 & 0.729 & 0.506 & 0.148 & 21.7 \\
jina + LR & 0.815 & 0.469 & 76.1 & 0.741 & 0.469 & 0.157 & 30.8 \\
Qwen3-Embedding + LR & 0.855 & 0.539 & 83.3 & 0.754 & 0.536 & 0.118 & 22.6 \\
Pooled probe & 0.868 & 0.595 & 79.5 & 0.767 & 0.534 & 0.138 & 30.1 \\
Fusion (A2 + TF-IDF) & 0.866 & 0.579 & 87.0 & 0.659 & 0.446 & 0.093 & 9.5 \\
\bottomrule
\end{tabular}}
\end{table}

\begin{table}[!htbp]
\centering
\caption{\textbf{Sensitivity analysis on the test set without patients discharged to hospice.} Pooled differences in AUROC (accuracy for the threshold contrast). ``Zero-shot'' denotes the prompt with matched wording and ``similar examples'' those retrieved with the original encoder unless stated otherwise.}
\label{tab:nohospice_test}
\small
\resizebox{\linewidth}{!}{%
\begin{tabular}{lrccc}
\toprule
Contrast & Difference & 95\% CI & Cancers $>0$ & $I^2$ (\%) \\
\midrule
Similar $-$ random examples (primary) & $+0.005$ & $-0.009$, $0.019$ & 8/11 & 30 \\
Random examples $-$ zero-shot & $-0.047$ & $-0.060$, $-0.034$ & 0/11 & 54 \\
Similar examples $-$ zero-shot & $-0.039$ & $-0.051$, $-0.027$ & 1/11 & 13 \\
Similar examples, true $-$ shuffled outcomes & $-0.007$ & $-0.017$, $0.004$ & 2/11 & 0 \\
Zero-shot with fitted threshold $-$ similar examples (accuracy) & $+0.026$ & $0.019$, $0.033$ & 11/11 & 61 \\
Similar examples (Qwen3-Embedding) $-$ random examples & $+0.023$ & $0.010$, $0.037$ & 9/11 & 28 \\
Similar examples (Qwen3-Embedding) $-$ zero-shot & $-0.021$ & $-0.032$, $-0.010$ & 1/11 & 0 \\
Zero-shot $-$ TF-IDF + LR & $+0.015$ & $0.002$, $0.028$ & 8/11 & 51 \\
Pooled probe $-$ zero-shot & $+0.010$ & $0.003$, $0.017$ & 8/11 & 41 \\
Pooled probe $-$ similar examples & $+0.049$ & $0.036$, $0.062$ & 11/11 & 17 \\
\bottomrule
\end{tabular}}
\end{table}

% ---------------------------------------------------------------------
\clearpage
\suppnote{Summary-based prompting}{supp:summary}

In its best-performing configuration, the original study showed the model LLM-written summaries of five similar patients with their outcomes, followed by a summary of the target patient\citep{anil2026dynamic}. We ran the closest analogue available to us; because our summaries were produced differently, these strategies are limited as a replication of that configuration. The same 8B model wrote the summaries and made the predictions, and no clinician reviewed or corrected the summaries, whereas in the original study GPT-4 Turbo wrote the summaries, about a third of which were sampled for clinician review, with discrepancies corrected if found. The 420-token generation limit was reached by 502 of 8{,}498 summaries of pool and development patients (5.9\%) and by 353 of 5{,}666 summaries of test patients (6.2\%), and these summaries are cut off mid-text. A zero-shot summary prompt has about 480 tokens and a five-shot summary prompt about 2{,}500, against about 2{,}700 and 16{,}000 tokens for the corresponding full-note prompts.

On the test set (Supplementary Table~\ref{tab:summary_test}; exploratory contrasts pooled across cancers), the AUROC with similar summarised examples differed from that with random summarised examples by $+0.007$ ($-0.002$ to $0.016$), an interval that lies within $\pm0.03$, as it did with full notes. Compared with the patient's summary alone, similar summarised examples changed AUROC by $-0.007$ ($-0.014$ to $0.000$) and were higher in 3 of 11 cancers; the corresponding loss with full notes was larger, at 0.034. Replacing notes by summaries raised the AUROC of the similar-example strategy by 0.028 ($0.018$ to $0.038$), whereas without examples the same replacement changed AUROC by $-0.002$ ($-0.010$ to $0.005$). Mean AUROC was 0.854 for the summary alone, 0.839 with random and 0.847 with similar summarised examples, against 0.858 for the patient's full note alone. Shorter inputs thus went with a smaller loss from examples, but even in the configuration closest to the best one of the original study, similar examples did not improve on a zero-shot prompt.

The development set showed the same attenuated pattern (Supplementary Table~\ref{tab:dev_pooled}). The interval for similar minus random summarised examples included zero ($+0.005$; $-0.006$ to $0.016$), similar summarised examples had a lower AUROC than the summary alone ($-0.014$; $-0.024$ to $-0.005$; lower in 10 of 11 cancers), and in every cancer the similar-example strategy did better with summaries than with full notes ($+0.030$; $0.017$ to $0.043$).

\begin{table}[!htbp]
\centering
\caption{\textbf{Summary-based strategies on the test set.} Top: pooled differences in AUROC (exploratory, not adjusted). Bottom: AUROC by cancer.}
\label{tab:summary_test}
\small
\begin{tabular}{lrccc}
\toprule
Contrast & $\dauc$ & 95\% CI & Cancers $>0$ & $I^2$ (\%) \\
\midrule
S4 summary similar $-$ S3 summary random & $+0.007$ & $-0.002$, $0.016$ & 9/11 & 0 \\
S4 summary similar $-$ S0 summary zero-shot & $-0.007$ & $-0.014$, $0.000$ & 3/11 & 0 \\
S4 summary similar $-$ A4 full-note similar & $+0.028$ & $0.018$, $0.038$ & 10/11 & 0 \\
S0 summary zero-shot $-$ A2 full-note zero-shot & $-0.002$ & $-0.010$, $0.005$ & 5/11 & 0 \\
\bottomrule
\end{tabular}

\vspace{0.8em}
\begin{tabular}{lrccc}
\toprule
Cancer & $N$ / deaths & S0 zero-shot & S3 random & S4 similar \\
\midrule
Lung & 1155 / 280 & 0.823 & 0.805 & 0.809 \\
Pancreas & 554 / 152 & 0.820 & 0.802 & 0.833 \\
Liver & 502 / 104 & 0.849 & 0.834 & 0.852 \\
Prostate & 689 / 59 & 0.880 & 0.877 & 0.868 \\
Colorectal & 573 / 59 & 0.883 & 0.872 & 0.879 \\
Non-Hodgkin lymphoma & 678 / 56 & 0.824 & 0.786 & 0.819 \\
Biliary tract & 177 / 43 & 0.884 & 0.879 & 0.850 \\
Breast & 620 / 40 & 0.925 & 0.917 & 0.921 \\
Bladder & 268 / 40 & 0.850 & 0.839 & 0.853 \\
Stomach & 209 / 40 & 0.847 & 0.822 & 0.827 \\
Brain (malignant) & 241 / 39 & 0.809 & 0.791 & 0.804 \\
\midrule
Mean over cancers & & 0.854 & 0.839 & 0.847 \\
\bottomrule
\end{tabular}
\end{table}

% ---------------------------------------------------------------------
\clearpage
\suppnote{Model scale: Qwen3-32B}{supp:scale}

We examined whether the conclusions depend on the size of the language model by running three strategies on the test set with AWQ-quantised Qwen3-32B: the patient's note alone with matched wording (A2), random examples (A3) and similar examples (A4). The prompts, retrieval, example library, truncation and score were the same as for Qwen3-8B. The model was run with vLLM on two RTX 3090 cards with tensor parallelism and a maximum context of 20{,}480 tokens, which is above the longest prompt (about 17{,}500 tokens). One brain cancer record had no valid score in the zero-shot strategy and was excluded from all three, which leaves 5{,}665 records. Qwen3-32B was not part of the development-set analyses. Its only earlier use was a pilot with 200 lung cancer patients (48 deaths), all of whom were later assigned to the development set. The pilot used an earlier prompt and an earlier procedure for reading the answer, which considered only the first generated position, so its numbers are not comparable with the final strategies; the AUROC was 0.839 for zero-shot, 0.807 for random and 0.767 for similar examples, the same ordering as for Qwen3-8B in the lung development set (Supplementary Table~\ref{tab:dev_auroc_prompt}).

On the test set, both kinds of example had a lower AUROC than the patient's note alone (Supplementary Table~\ref{tab:scale}). Similar examples differed from the note alone by $-0.022$ ($-0.030$ to $-0.013$; Holm-adjusted $p<0.001$ within this family of three contrasts) and were lower in 10 of 11 cancers, brain being the exception ($+0.006$). Random examples differed by $-0.013$ ($-0.020$ to $-0.005$; adjusted $p=0.001$) and were lower in 10 of 11 cancers, prostate being the exception ($+0.002$). The losses were about two thirds (similar) and one third (random) of those seen with Qwen3-8B. The pooled difference between similar and random examples was $-0.009$ ($-0.017$ to $-0.001$; adjusted $p=0.034$; 4 of 11 cancers positive; $I^2=0$\%). This interval lies inside the $\pm0.03$ margin and also excludes zero, on the side of random examples, so that under our decision rule the two strategies are both equivalent and different; with Qwen3-8B only the equivalence criterion was met. Both models give the same answer to the primary question: similarity gave no benefit. We do not read the small negative difference as evidence that similar examples are harmful, because its sign differs between the two models.

In a comparison that was not in the frozen plan (post hoc, unadjusted), Qwen3-32B had a higher AUROC than Qwen3-8B with every strategy: $+0.014$ ($0.008$ to $0.020$) for the note alone, $+0.041$ ($0.031$ to $0.051$) with random and $+0.026$ ($0.016$ to $0.036$) with similar examples. The gain was thus largest where the smaller model had lost most, and the zero-shot prompt still had the highest AUROC of the three. The comparison does not isolate the effect of parameter count from that of quantisation.

As with Qwen3-8B, adding examples moved accuracy at the default threshold and AUROC in opposite directions. The model predicted death for 33.7\% of patients from the note alone, 21.2\% with random and 15.8\% with similar examples, against an observed mortality of 16.4\%; accuracy rose from 77.0\% to 83.1\% and 85.0\%, and the Brier score fell from 0.181 to 0.140 and 0.123 (means over the 11 cancers). After the exclusion of patients discharged to hospice, the three contrasts kept their direction: $-0.012$ for similar minus random examples, $-0.014$ for random examples minus the note alone and $-0.025$ for similar examples minus the note alone. The probe, the alternative retriever and the analyses of neighbours and hidden states were not repeated with Qwen3-32B.

\begin{table}[!htbp]
\centering
\caption{\textbf{Qwen3-32B on the test set, with Qwen3-8B for comparison.} AUROC is the unweighted mean over the 11 cancers, and differences are pooled by inverse variance. The last three rows are post hoc.}
\label{tab:scale}
\small
\begin{tabular}{lcccc}
\toprule
 & \multicolumn{2}{c}{Qwen3-8B} & \multicolumn{2}{c}{Qwen3-32B} \\
\cmidrule(lr){2-3}\cmidrule(lr){4-5}
 & Estimate & 95\% CI & Estimate & 95\% CI \\
\midrule
A2 zero-shot, mean AUROC & 0.858 & & 0.871 & \\
A3 random, mean AUROC & 0.809 & & 0.855 & \\
A4 similar, mean AUROC & 0.818 & & 0.848 & \\
\midrule
A4 $-$ A3 (similar $-$ random) & $+0.006$ & $-0.006$, $0.018$ & $-0.009$ & $-0.017$, $-0.001$ \\
A4 $-$ A2 (similar $-$ zero-shot) & $-0.034$ & $-0.045$, $-0.024$ & $-0.022$ & $-0.030$, $-0.013$ \\
A3 $-$ A2 (random $-$ zero-shot) & $-0.043$ & $-0.054$, $-0.032$ & $-0.013$ & $-0.020$, $-0.005$ \\
\midrule
 & \multicolumn{4}{c}{Qwen3-32B $-$ Qwen3-8B (post hoc)} \\
A2 zero-shot & \multicolumn{2}{c}{} & $+0.014$ & $0.008$, $0.020$ \\
A3 random & \multicolumn{2}{c}{} & $+0.041$ & $0.031$, $0.051$ \\
A4 similar & \multicolumn{2}{c}{} & $+0.026$ & $0.016$, $0.036$ \\
\bottomrule
\end{tabular}
\end{table}

\paragraph*{Compute resources for the Qwen3-8B analyses}
Development-set runs used one RTX 5090 card, and test-set runs a server with two RTX 3090 cards, on which the full-note strategies ran with tensor parallelism across both cards and the summary-based strategies and the Qwen3-Embedding retrieval strategy (R3) on a single card. Scores from the two set-ups agreed closely but not exactly: on 120 lung cancer patients from the development set, for the zero-shot prompt with matched wording and for similar examples, their correlation was at least 0.999, the largest difference for any single patient was 0.09 and the paired AUROCs differed by at most 0.003.

% ---------------------------------------------------------------------
\clearpage
\suppnote{Retrieved neighbours, influence of examples and hidden states}{supp:mech}

This note gives the tables that underlie Fig.~\ref{fig:mechanism} and additional detail on the exploratory analyses of neighbours and hidden states. Unless marked otherwise, values come from the development set, in which the library is the pool.

\paragraph{Retrieved neighbours.}
In the development set, as in the test set, retrieved patients shared the target's admission type and discharge location more often than random patients did, and the neighbour vote reached an AUROC of 0.72 to 0.74, against 0.50 for random patients (Supplementary Table~\ref{tab:neighbours_dev}); the three retrievers differed little on these measures. The two encoders spread patients out differently. In the jina embedding space all patients lay close together: mean cosine similarity to the target was 0.972 for its jina neighbours and 0.935 for random patients. Qwen3-Embedding separated them more (0.789 for its own neighbours against 0.626 for random patients). This difference was accompanied by only a small difference in neighbour-vote AUROC, and neighbours chosen by one encoder were also closer to the target than random patients were in the other encoder's space. The original study assessed its neighbours by accuracy instead: in its five-shot summary setting, the majority outcome of the retrieved neighbours was correct for 65\% of glioma and 74\% of breast cancer patients\citep{anil2026dynamic}, against 61.5\% and 77\% for always predicting the majority class (computed from its Table 1).

\begin{table}[!htbp]
\centering
\caption{\textbf{Properties of the five library patients selected for each target in the development set.} Values are unweighted means over the 11 cancers, by selection method. ``Neighbour-vote AUROC'' uses the proportion of deaths among the five as the risk score, and ``outcome agreement'' is the proportion of selected patients with the same 90-day outcome as the target. Cosine similarities are between the target and the selected patients, in each of the two embedding spaces. The last two rows count targets, out of 3{,}682, whose five selected patients share one outcome.}
\label{tab:neighbours_dev}
\small
\begin{tabular}{lcccc}
\toprule
 & Random & TF-IDF & jina (original) & Qwen3-Embedding \\
\midrule
Neighbour-vote AUROC            & 0.499 & 0.717 & 0.729 & 0.742 \\
Outcome agreement               & 0.735 & 0.781 & 0.772 & 0.794 \\
Same admission type             & 0.230 & 0.452 & 0.420 & 0.425 \\
Same discharge location         & 0.272 & 0.400 & 0.370 & 0.403 \\
Same sex                        & 0.617 & 0.693 & 0.673 & 0.711 \\
Absolute age difference (years) & 14.6  & 13.0  & 13.0  & 12.3 \\
Cosine similarity, jina space   & 0.935 & 0.961 & 0.972 & 0.965 \\
Cosine similarity, Qwen3-Embedding space & 0.626 & 0.729 & 0.752 & 0.789 \\
\midrule
Targets with five survivors     & 1{,}782 & 2{,}022 & 1{,}871 & 2{,}114 \\
Targets with five deaths        & 1 & 14 & 19 & 24 \\
\bottomrule
\end{tabular}
\end{table}

\paragraph{Training of the probe.}
The layer and the regularisation parameter of the probe were chosen by AUROC on out-of-fold predictions from five-fold cross-validation within the training set, with folds stratified by outcome at the record level and standardisation fitted on the training records of each fold; for the pooled probe, records from the same patient could appear in different internal folds. Two versions were evaluated on the development set, one probe per cancer and one pooled probe trained on the pools of all 11 cancers together. For the pooled probe, cross-validation selected the final layer and $C=10^{-3}$ (cross-validated AUROC 0.877). The per-cancer probes were unstable in small cohorts (Supplementary Note~\ref{supp:dev}), and only the pooled probe was evaluated on the test set.

\paragraph{Labels required by the probe.}
We trained the pooled probe on subsets of the labelled pool to see how its performance depended on the number of labels (Supplementary Table~\ref{tab:learning}). With 100 to 250 labelled training records, its mean AUROC was below that of the zero-shot answer; the two were about equal at 500 records, and the probe gained another 0.015 by 4{,}470 records.

\begin{table}[!htbp]
\centering
\caption{\textbf{Mean development-set AUROC of the pooled probe as a function of the number of labelled training records.} The zero-shot prompt (A2), which uses no labels, has a mean AUROC of 0.846.}
\label{tab:learning}
\small
\begin{tabular}{lcccccc}
\toprule
Training records & 100 & 250 & 500 & 1{,}000 & 2{,}000 & 4{,}470 \\
\midrule
Pooled probe, mean AUROC & 0.836 & 0.840 & 0.845 & 0.855 & 0.858 & 0.860 \\
\bottomrule
\end{tabular}
\end{table}

\paragraph{Influence of the examples on the answer.}
In every strategy, the few-shot answer followed the zero-shot reading incompletely ($\beta_{\mathrm{zs}}$ between 0.55 and 0.77) and moved with the number of deaths shown (positive $\beta_d$; Supplementary Table~\ref{tab:anchor}). Two controls gave coefficients similar to those for similar examples with true outcomes (A4): shuffling the outcomes among the same five similar patients (A5), and showing random patients with the same number of deaths as the A4 examples (R1). This similarity suggests a role for the count of deaths, and the choice of retriever made little difference. Both controls are compatible with the account given in the main text. Shuffling the outcomes keeps the death count fixed, and the AUROC of A5 was close to that of A4. With the death count matched, random texts gave a higher AUROC than similar texts ($+0.019$, in all 11 cancers; Supplementary Table~\ref{tab:dev_pooled}), and the model's own reading kept slightly more weight ($\beta_{\mathrm{zs}}$ 0.59 against 0.56). The regression gave nearly the same coefficients on the test set. The strategies with death-matched random examples (R1) and with TF-IDF-retrieved examples (R2) were run on the development set only, although the outcome votes of the TF-IDF neighbours were also evaluated on the test set (Extended Data Table~\ref{tab:neighbours_test}).

The meaning of $\beta_d$ differs between strategies. In every strategy except random examples (A3), the death count comes from the target's neighbours and is correlated with the target's risk; part of $\beta_d$ may then reflect legitimate use of that information, or risk-related features of the target that the zero-shot log-odds capture imperfectly. In A3 the count is independent of the target by construction, and its positive $\beta_d$ is therefore the cleanest evidence in this analysis that outcomes shown in the prompt shift the answer irrespective of the target. Because the count has a different distribution in A3, the size of its standardised coefficient cannot be compared with that of the other strategies, and we treat the table as descriptive.

\begin{table}[!htbp]
\centering
\caption{\textbf{Standardised coefficients of Eq.~\eqref{eq:anchor}, pooled over 11 cancers (95\% CI).} Top: development set; ``Texts'' and ``Death count'' describe what the five examples of each strategy consist of, and in A5 and R1 the death count equals that of the jina neighbours by construction. Bottom: test set.}
\label{tab:anchor}
\small
\resizebox{\linewidth}{!}{%
\begin{tabular}{llllcc}
\toprule
Strategy & Texts & Death count & Outcome--text link & $\beta_{\mathrm{zs}}$ (own reading) & $\beta_d$ (deaths shown) \\
\midrule
A3 & random  & random          & true     & 0.770 (0.755, 0.785) & 0.167 (0.145, 0.188) \\
R1 & random  & as neighbours   & true     & 0.594 (0.572, 0.615) & 0.374 (0.351, 0.397) \\
A4 & similar (jina) & neighbours & true   & 0.556 (0.534, 0.578) & 0.398 (0.374, 0.421) \\
A5 & similar (jina) & neighbours & shuffled & 0.552 (0.530, 0.573) & 0.423 (0.400, 0.445) \\
R2 & similar (TF-IDF) & neighbours & true & 0.584 (0.562, 0.606) & 0.394 (0.371, 0.417) \\
R3 & similar (Qwen3-Emb.) & neighbours & true & 0.585 (0.563, 0.607) & 0.412 (0.389, 0.435) \\
\bottomrule
\end{tabular}}

\vspace{0.8em}
\begin{tabular}{lcc}
\toprule
Strategy (test set) & $\beta_{\mathrm{zs}}$ (own reading) & $\beta_d$ (deaths shown) \\
\midrule
A3 random & 0.775 (0.763, 0.787) & 0.173 (0.156, 0.190) \\
A4 similar (jina) & 0.547 (0.529, 0.564) & 0.428 (0.410, 0.447) \\
A5 similar, shuffled & 0.543 (0.525, 0.560) & 0.441 (0.423, 0.458) \\
R3 similar (Qwen3-Emb.) & 0.569 (0.552, 0.587) & 0.409 (0.391, 0.427) \\
\bottomrule
\end{tabular}
\end{table}

\paragraph{Hidden states without and with examples.}
Hidden states were taken after the final normalisation layer at the last prompt token, and the answer direction $u$ of Eq.~\eqref{eq:readout} was built from the output-embedding rows of the tokens \texttt{NEG} and \texttt{POS}. Without examples, the projection of the state on $u$ had a correlation of 0.988 (mean over cancers) with the zero-shot log-odds read from the model and the same mean AUROC (0.846). The model first emits an opening quotation mark, and the score is read one position later; the information behind the answer is therefore already largely present in the state from which that first token is generated. When the component of each hidden state along $u$ was removed before training, the pooled probe still reached 0.860, so that the probe is more than a recalibration of the verbal answer. For the states with examples, pool patients retrieved their five similar examples from the rest of the pool, and the pooled probe was trained with the same training records and regularisation as before.

Supplementary Table~\ref{tab:fs_hidden} compares readouts of the states without and with five similar examples. The probe trained on few-shot states (mean AUROC 0.845) was 0.049 above the model's own few-shot answer (95\% CI $0.035$ to $0.064$), without a detectable difference from its zero-shot answer ($-0.001$; $-0.015$ to $0.012$), and 0.015 below the probe trained on zero-shot states ($0.003$ to $0.028$). Both probes were fitted with the same labels, so their comparison concerns the risk discrimination achieved by these linear readouts. The probe trained on zero-shot states transferred poorly to few-shot states (AUROC 0.755 at the final layer and 0.79 to 0.80 at layers 24 to 32). As in the zero-shot case, projecting the few-shot state on the answer direction closely reproduced the few-shot answer (correlation 0.983 with its log-odds). The AUROC of the few-shot probe cannot be attributed to the target note alone, because the few-shot state also contains information about the examples, including their outcomes, which predict the target's outcome to some degree. The comparison therefore does not quantify separate contributions from the representation and the readout, and it covers the original retriever and the final layers only.

\begin{table}[!htbp]
\centering
\caption{\textbf{Mean development-set AUROC over 11 cancers for different readouts of the final-layer hidden state,} without examples (A2 prompt) and with five similar examples (A4 prompt). Probes were trained on 4{,}470 labelled pool records.}
\label{tab:fs_hidden}
\small
\begin{tabular}{lcc}
\toprule
Readout & Zero-shot state & Few-shot state \\
\midrule
Model's answer (score of Eq.~\eqref{eq:score}) & 0.846 & 0.795 \\
Projection on the answer direction $u$            & 0.846 & 0.811 \\
Probe trained on zero-shot states                 & 0.860 & 0.755 \\
Probe trained on the same kind of state           & 0.860 & 0.845 \\
\bottomrule
\end{tabular}
\end{table}

% ---------------------------------------------------------------------
\clearpage
\suppnote{Fusion of the zero-shot score with a text classifier}{supp:fusion}

In this exploratory analysis, we asked whether the zero-shot score and a bag-of-words classifier carry complementary information. We combined them with a logistic regression on two inputs: the logit of the zero-shot (A2) score and the logit of the TF-IDF + LR score. To keep the fit out of sample, the TF-IDF classifier used here was trained on the pool only; the two fusion weights were then fitted per cancer on the development set and applied unchanged to the test set.

On the test set, the fused score had a mean AUROC of 0.866. In pooled differences, it was 0.032 above its own TF-IDF component ($0.025$ to $0.038$; higher in all 11 cancers) and 0.024 above the TF-IDF classifier trained on the pool plus the development set ($0.017$ to $0.031$; Supplementary Table~\ref{tab:fusion_test}). The zero-shot score thus adds information to a text classifier. Evidence for the reverse is descriptive, because no contrast between the fused score and the zero-shot score alone was planned: their mean AUROCs were 0.866 and 0.858, respectively, and the fused score was the higher of the two in 8 of 11 cancers.

\begin{table}[!htbp]
\centering
\caption{\textbf{Fusion on the test set (exploratory).} Top: pooled differences in AUROC. Bottom: AUROC by cancer.}
\label{tab:fusion_test}
\small
\begin{tabular}{lrccc}
\toprule
Contrast & $\dauc$ & 95\% CI & Cancers $>0$ & $I^2$ (\%) \\
\midrule
Fusion $-$ TF-IDF (pool), its own component & $+0.032$ & $0.025$, $0.038$ & 11/11 & 73 \\
Fusion $-$ TF-IDF (pool + dev) & $+0.024$ & $0.017$, $0.031$ & 10/11 & 60 \\
\bottomrule
\end{tabular}

\vspace{0.8em}
\resizebox{\linewidth}{!}{%
\begin{tabular}{lrcccc}
\toprule
Cancer & $N$ / deaths & A2 zero-shot & TF-IDF (pool) & TF-IDF (pool + dev) & Fusion \\
\midrule
Lung & 1155 / 280 & 0.825 & 0.831 & 0.832 & 0.848 \\
Pancreas & 554 / 152 & 0.839 & 0.781 & 0.808 & 0.852 \\
Liver & 502 / 104 & 0.844 & 0.814 & 0.852 & 0.856 \\
Prostate & 689 / 59 & 0.878 & 0.820 & 0.836 & 0.869 \\
Colorectal & 573 / 59 & 0.872 & 0.852 & 0.863 & 0.887 \\
Non-Hodgkin lymphoma & 678 / 56 & 0.851 & 0.751 & 0.748 & 0.835 \\
Biliary tract & 177 / 43 & 0.858 & 0.839 & 0.838 & 0.890 \\
Breast & 620 / 40 & 0.928 & 0.908 & 0.909 & 0.936 \\
Bladder & 268 / 40 & 0.869 & 0.759 & 0.791 & 0.857 \\
Stomach & 209 / 40 & 0.876 & 0.807 & 0.822 & 0.880 \\
Brain (malignant) & 241 / 39 & 0.798 & 0.789 & 0.837 & 0.813 \\
\midrule
Mean over cancers & & 0.858 & 0.814 & 0.831 & 0.866 \\
\bottomrule
\end{tabular}}
\end{table}

% ---------------------------------------------------------------------
\clearpage
\suppnote{Development-set results}{supp:dev}

All results in this note come from the development set (3{,}682 patient--cancer records, 546 deaths), in which examples and training data were drawn from the pool only. They were seen before the analysis plan was frozen and informed the design, and they are not confirmatory. No response failed to parse. For each cancer, the paired difference between two strategies was calculated with a 95\% paired-bootstrap interval (2{,}000 resamples); the 11 differences were pooled by inverse variance with a normal approximation for the pooled interval. No adjustment was made for multiple comparisons, and patients who appear in more than one cohort were not resampled jointly, so that pooled intervals may be somewhat too narrow.

\begin{figure}[!htbp]
\centering
\includegraphics[width=\linewidth]{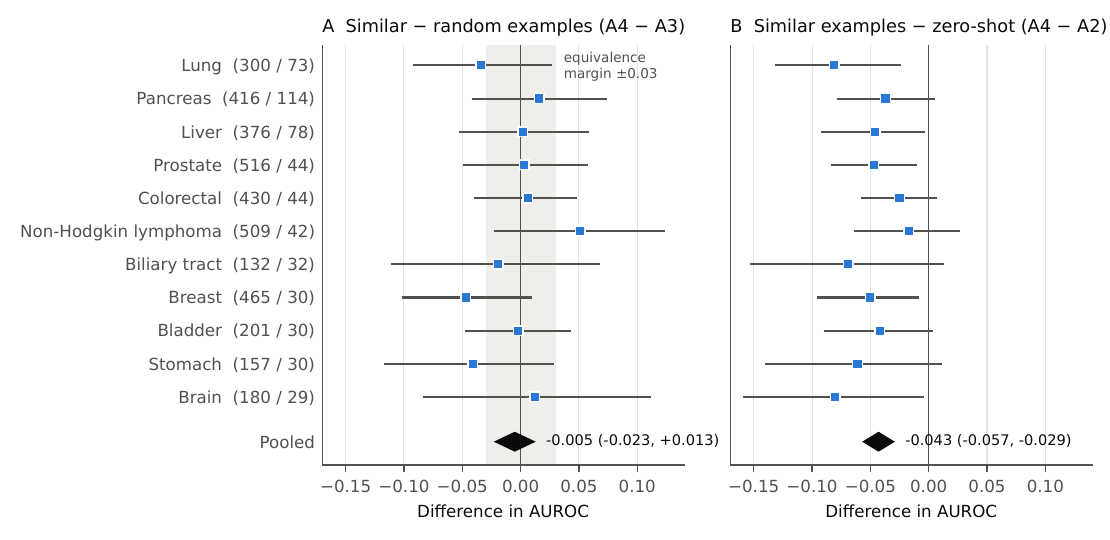}
\caption{\textbf{Per-cancer and pooled differences in AUROC in the development set,} with 95\% paired-bootstrap intervals; numbers in parentheses are patients / deaths. (A) Similar minus random examples (A4 $-$ A3); the shaded band is the equivalence margin of $\pm0.03$. (B) Similar examples minus the patient's note alone with matched wording (A4 $-$ A2). Diamonds are inverse-variance pooled estimates.}
\label{fig:forest_dev}
\end{figure}

\begin{figure}[!htbp]
\centering
\includegraphics[width=\linewidth]{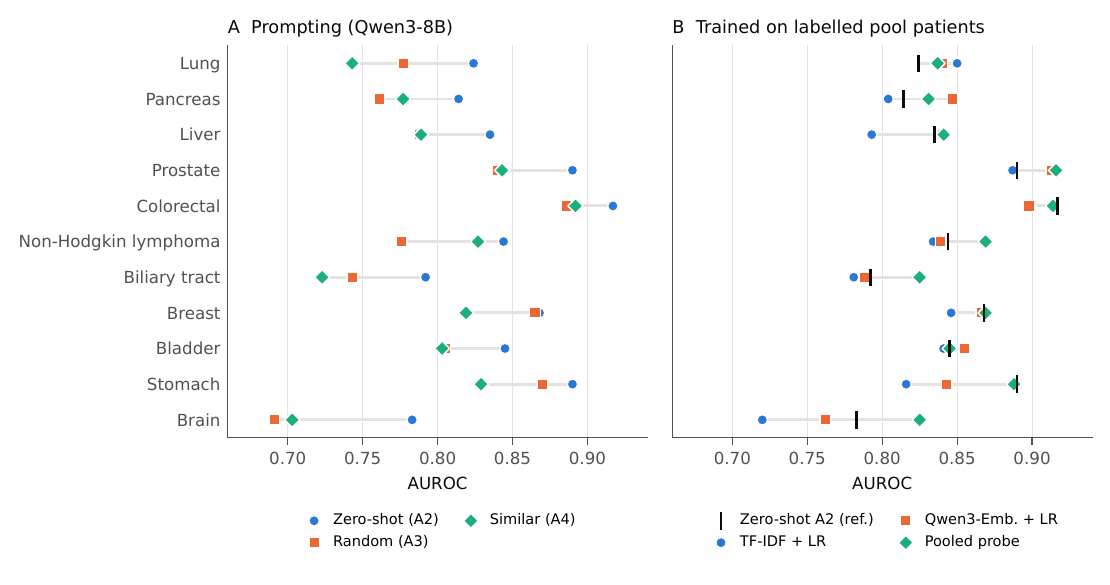}
\caption{\textbf{Development-set AUROC by cancer.} (A) Prompting with Qwen3-8B: the patient's note alone with matched wording (A2), random examples (A3) and similar examples (A4). (B) Classifiers trained on labelled pool records: TF-IDF + LR, Qwen3-Embedding + LR and the pooled probe; the black tick marks the zero-shot prompt (A2). The methods in (B) use labelled training data and serve as reference points, not as like-for-like alternatives to prompting.}
\label{fig:dots_dev}
\end{figure}

\paragraph{Similar examples, random examples and the note alone.}
The pooled AUROC difference between similar and random examples was $-0.005$ ($-0.023$ to $0.013$; Supplementary Table~\ref{tab:dev_pooled}). Its interval lay inside $\pm0.03$, and no single cancer had an interval excluding zero (Supplementary Table~\ref{tab:dev_contrasts}, Supplementary Fig.~\ref{fig:forest_dev}). An earlier analysis of the first three cancers alone had given $-0.033$ ($-0.071$ to $0.005$); most of that apparent disadvantage of similar examples did not persist with 11 cancers. Both similar and random examples gave a lower AUROC than the patient's note alone with matched wording in every cancer (Supplementary Fig.~\ref{fig:dots_dev}), and for similar examples five of the 11 cancer-specific intervals excluded zero.

\paragraph{Number of deaths shown, shuffled outcomes and choice of retriever.}
When the number of deaths among the examples was matched, similar examples (A4) had a lower AUROC than the death-matched random examples (R1) in all 11 cancers, with a pooled difference of $-0.019$ ($-0.034$ to $-0.003$). This difference had the same direction as the primary contrast in this set but was larger; it is an unadjusted secondary comparison. The comparison between true and shuffled outcomes is diluted, because for 1{,}890 of the 3{,}682 development records (51\%; 1{,}871 with five survivors and 19 with five deaths) the five neighbours shared one outcome and the two prompts were identical. Similar examples selected with Qwen3-Embedding (R3) gave a higher AUROC than similar examples selected with the original retriever. Their difference from random examples, $+0.013$ ($-0.003$ to $0.030$), is inconclusive because the interval includes zero and reaches the equivalence margin, and R3 remained below the patient's note alone in every cancer.

\paragraph{Probes, variation across cancers and hospice discharges.}
The pooled probe was trained on 4{,}470 patient--cancer records from the pool, including 757 deaths; cross-validation in the pool selected layer 36 and $C=10^{-3}$ (cross-validated AUROC 0.877). The probe trained separately for each cancer was unstable in small cohorts: for brain tumours, cross-validation selected layer 18 and the development AUROC was 0.559. Heterogeneity across cancers was low for the main contrasts ($I^2=0$\% for most and 20--24\% for two) and was higher only for the per-cancer probe and for the comparison of the two embedding classifiers. After the exclusion of patients discharged to hospice, the AUROC of the note alone fell in every cancer, similar examples stayed below the note alone in all 11, and the pooled contrasts changed by at most 0.007 (last block of Supplementary Table~\ref{tab:dev_pooled}).

\paragraph{Accuracy and the proportion of patients predicted to die.}
At the fixed threshold of 0.5, accuracy for the note alone was higher with the matched wording (A2) than with the original wording (A1) in every cancer, and both random (A3) and similar examples (A4) were more accurate than A2 in every cancer except bladder cancer (Supplementary Table~\ref{tab:dev_accuracy}). The change of wording from A1 to A2 raised accuracy by more than the addition of examples did in 10 of 11 cancers, pancreatic cancer being the exception. Over the same steps, the model predicted death less often (Supplementary Table~\ref{tab:dev_predrate}). Under A1, A2 and A4, respectively, the proportion of patients predicted to die was 82.7\%, 60.0\% and 31.0\% for lung cancer, 35.9\%, 16.3\% and 6.0\% for breast cancer, and 72.2\%, 25.6\% and 10.0\% for brain tumours. Under A2, the model predicted death for 86\% of patients with pancreatic cancer (death rate 27\%) and 48\% of patients with liver cancer (21\%); with random or similar examples, the predicted proportion came closer to the observed rate in these cohorts (36--40\% and 20--25\%, respectively). When the 0.5 threshold for A2 was replaced by a cross-fitted threshold for each cancer, A2 was more accurate than the better of A3 and A4 in 10 of 11 cancers and tied in one (non-Hodgkin lymphoma), and it was more accurate than predicting survival for every patient in 10 of 11, falling slightly below it for breast cancer.

\begin{table}[!htbp]
\centering
\caption{\textbf{Pooled differences in AUROC in the development set (Qwen3-8B).} ``Cancers $>0$'' is the number of cancers with a positive difference and $I^2$ the heterogeneity across cancers (for some rows approximated from the widths of the cancer-specific intervals); ``--'', not computed.}
\label{tab:dev_pooled}
\small
\begin{tabular}{lrccc}
\toprule
Contrast & $\dauc$ & 95\% CI & Cancers $>0$ & $I^2$ (\%) \\
\midrule
\multicolumn{5}{l}{\emph{Do examples help?}} \\
A4 similar $-$ A3 random & $-0.005$ & $-0.023$, $0.013$ & 6/11 & 0 \\
A4 similar $-$ A2 zero-shot & $-0.043$ & $-0.057$, $-0.029$ & 0/11 & 0 \\
A3 random $-$ A2 zero-shot & $-0.040$ & $-0.054$, $-0.027$ & 0/11 & 0 \\
A4 similar $-$ A5 outcomes shuffled & $-0.011$ & $-0.024$, $0.002$ & 3/11 & 24 \\
A4 similar $-$ R1 random, death-matched & $-0.019$ & $-0.034$, $-0.003$ & 0/11 & 0 \\
\midrule
\multicolumn{5}{l}{\emph{Retriever}} \\
R2 (TF-IDF) $-$ A4 (jina) & $+0.005$ & $-0.011$, $0.022$ & 5/11 & 0 \\
R3 (Qwen3-Embedding) $-$ A4 (jina) & $+0.018$ & $0.002$, $0.034$ & 9/11 & 0 \\
R3 $-$ R2 & $+0.012$ & $-0.003$, $0.028$ & 8/11 & -- \\
R3 $-$ A3 random & $+0.013$ & $-0.003$, $0.030$ & 7/11 & 20 \\
R3 $-$ A2 zero-shot & $-0.028$ & $-0.042$, $-0.013$ & 0/11 & 0 \\
\midrule
\multicolumn{5}{l}{\emph{Summaries}} \\
S4 summary similar $-$ A4 full-note similar & $+0.030$ & $0.017$, $0.043$ & 11/11 & 0 \\
S4 $-$ S3 (summary: similar $-$ random) & $+0.005$ & $-0.006$, $0.016$ & 6/11 & 0 \\
S4 $-$ S0 (summary: similar $-$ zero-shot) & $-0.014$ & $-0.024$, $-0.005$ & 1/11 & 0 \\
\midrule
\multicolumn{5}{l}{\emph{Supervised comparators and probe}} \\
A2 zero-shot $-$ TF-IDF + LR & $+0.014$ & $-0.002$, $0.031$ & 10/11 & 0 \\
A4 similar $-$ TF-IDF + LR & $-0.029$ & $-0.048$, $-0.010$ & 1/11 & 1 \\
Qwen3-Embedding + LR $-$ TF-IDF + LR & $+0.017$ & $0.005$, $0.030$ & 9/11 & 0 \\
Qwen3-Embedding + LR $-$ jina + LR & $+0.041$ & $0.028$, $0.054$ & 11/11 & 54 \\
Fusion (A2 + TF-IDF) $-$ TF-IDF + LR & $+0.021$ & $0.010$, $0.031$ & 11/11 & 0 \\
Pooled probe $-$ A2 zero-shot & $+0.010$ & $0.001$, $0.018$ & 7/11 & 0 \\
Per-cancer probe $-$ A2 zero-shot & $+0.001$ & $-0.008$, $0.010$ & 5/11 & 45 \\
Pooled probe $-$ A4 similar & $+0.055$ & $0.040$, $0.070$ & 11/11 & 1 \\
Pooled probe $-$ TF-IDF + LR & $+0.027$ & $0.013$, $0.042$ & 10/11 & 14 \\
Pooled probe $-$ Qwen3-Embedding + LR & $+0.007$ & $-0.006$, $0.020$ & 8/11 & 0 \\
\midrule
\multicolumn{5}{l}{\emph{Without hospice discharges}} \\
A4 similar $-$ A3 random & $-0.007$ & $-0.028$, $0.014$ & 5/11 & 0 \\
A4 similar $-$ A2 zero-shot & $-0.050$ & $-0.066$, $-0.033$ & 0/11 & 0 \\
A3 random $-$ A2 zero-shot & $-0.045$ & $-0.061$, $-0.029$ & 1/11 & 0 \\
\bottomrule
\end{tabular}
\end{table}

\begin{table}[!htbp]
\centering
\caption{\textbf{Development-set AUROC by cancer for the prompting strategies (Qwen3-8B).}}
\label{tab:dev_auroc_prompt}
\small
\resizebox{\linewidth}{!}{%
\begin{tabular}{lrcccccccc}
\toprule
Cancer & $N$ / deaths & A1 & A2 & A3 & A4 & A5 & R3 & S0 & S4 \\
\midrule
Lung                 & 300 / 73  & 0.801 & 0.824 & 0.777 & 0.743 & 0.804 & 0.763 & 0.788 & 0.756 \\
Breast               & 465 / 30  & 0.870 & 0.868 & 0.865 & 0.819 & 0.824 & 0.829 & 0.874 & 0.859 \\
Brain (malignant)    & 180 / 29  & 0.747 & 0.783 & 0.691 & 0.703 & 0.725 & 0.757 & 0.795 & 0.775 \\
Stomach              & 157 / 30  & 0.864 & 0.890 & 0.870 & 0.829 & 0.858 & 0.851 & 0.878 & 0.869 \\
Bladder              & 201 / 30  & 0.853 & 0.845 & 0.805 & 0.803 & 0.854 & 0.824 & 0.855 & 0.824 \\
Biliary tract        & 132 / 32  & 0.747 & 0.792 & 0.743 & 0.723 & 0.779 & 0.765 & 0.812 & 0.790 \\
Liver                & 376 / 78  & 0.790 & 0.835 & 0.788 & 0.789 & 0.763 & 0.832 & 0.846 & 0.819 \\
Pancreas             & 416 / 114 & 0.718 & 0.814 & 0.761 & 0.777 & 0.794 & 0.808 & 0.823 & 0.821 \\
Colorectal           & 430 / 44  & 0.885 & 0.917 & 0.886 & 0.892 & 0.899 & 0.883 & 0.909 & 0.897 \\
Non-Hodgkin lymphoma & 509 / 42  & 0.830 & 0.844 & 0.776 & 0.827 & 0.825 & 0.798 & 0.852 & 0.868 \\
Prostate             & 516 / 44  & 0.894 & 0.890 & 0.840 & 0.843 & 0.836 & 0.881 & 0.893 & 0.881 \\
\midrule
Mean over cancers    &           & 0.818 & 0.846 & 0.800 & 0.795 & 0.815 & 0.817 & 0.848 & 0.833 \\
\bottomrule
\end{tabular}}
\end{table}

\begin{table}[!htbp]
\centering
\caption{\textbf{Development-set AUROC by cancer for the supervised comparators,} with the patient's note alone (A2) for reference. The pooled probe uses labelled training data from all 11 pools.}
\label{tab:dev_auroc_sup}
\small
\begin{tabular}{lccccc}
\toprule
Cancer & A2 & TF-IDF + LR & Qwen3-Emb.\ + LR & Fusion & Pooled probe \\
\midrule
Lung                 & 0.824 & 0.850 & 0.840 & 0.852 & 0.837 \\
Breast               & 0.868 & 0.846 & 0.866 & 0.860 & 0.869 \\
Brain (malignant)    & 0.783 & 0.720 & 0.762 & 0.781 & 0.825 \\
Stomach              & 0.890 & 0.816 & 0.843 & 0.877 & 0.888 \\
Bladder              & 0.845 & 0.841 & 0.855 & 0.856 & 0.845 \\
Biliary tract        & 0.792 & 0.781 & 0.788 & 0.788 & 0.825 \\
Liver                & 0.835 & 0.793 & 0.840 & 0.839 & 0.841 \\
Pancreas             & 0.814 & 0.804 & 0.847 & 0.835 & 0.831 \\
Colorectal           & 0.917 & 0.899 & 0.898 & 0.927 & 0.914 \\
Non-Hodgkin lymphoma & 0.844 & 0.834 & 0.839 & 0.862 & 0.869 \\
Prostate             & 0.890 & 0.887 & 0.913 & 0.901 & 0.916 \\
\midrule
Mean over cancers    & 0.846 & 0.825 & 0.845 & 0.853 & 0.860 \\
\bottomrule
\end{tabular}
\end{table}

\begin{table}[!htbp]
\centering
\caption{\textbf{Cancer-specific differences in AUROC for the main contrasts in the development set,} with 95\% paired-bootstrap intervals.}
\label{tab:dev_contrasts}
\small
\resizebox{\linewidth}{!}{%
\begin{tabular}{lrcrcrc}
\toprule
 & \multicolumn{2}{c}{A4 similar $-$ A3 random} & \multicolumn{2}{c}{A4 similar $-$ A2 zero-shot} & \multicolumn{2}{c}{A3 random $-$ A2 zero-shot} \\
\cmidrule(lr){2-3}\cmidrule(lr){4-5}\cmidrule(lr){6-7}
Cancer & $\dauc$ & 95\% CI & $\dauc$ & 95\% CI & $\dauc$ & 95\% CI \\
\midrule
Lung                 & $-0.034$ & $-0.092$, $0.027$ & $-0.081$ & $-0.132$, $-0.024$ & $-0.047$ & $-0.097$, $0.006$ \\
Breast               & $-0.047$ & $-0.102$, $0.010$ & $-0.050$ & $-0.096$, $-0.008$ & $-0.003$ & $-0.045$, $0.042$ \\
Brain (malignant)    & $+0.012$ & $-0.084$, $0.112$ & $-0.080$ & $-0.159$, $-0.004$ & $-0.092$ & $-0.168$, $-0.025$ \\
Stomach              & $-0.041$ & $-0.117$, $0.029$ & $-0.061$ & $-0.140$, $0.011$ & $-0.020$ & $-0.070$, $0.033$ \\
Bladder              & $-0.002$ & $-0.048$, $0.043$ & $-0.042$ & $-0.090$, $0.004$ & $-0.040$ & $-0.075$, $-0.008$ \\
Biliary tract        & $-0.019$ & $-0.111$, $0.068$ & $-0.069$ & $-0.153$, $0.013$ & $-0.050$ & $-0.121$, $0.014$ \\
Liver                & $+0.002$ & $-0.053$, $0.059$ & $-0.046$ & $-0.092$, $-0.003$ & $-0.048$ & $-0.090$, $-0.004$ \\
Pancreas             & $+0.016$ & $-0.042$, $0.074$ & $-0.037$ & $-0.079$, $0.005$ & $-0.053$ & $-0.106$, $-0.004$ \\
Colorectal           & $+0.006$ & $-0.040$, $0.048$ & $-0.025$ & $-0.058$, $0.007$ & $-0.031$ & $-0.062$, $0.000$ \\
Non-Hodgkin lymphoma & $+0.051$ & $-0.023$, $0.124$ & $-0.017$ & $-0.064$, $0.027$ & $-0.069$ & $-0.124$, $-0.019$ \\
Prostate             & $+0.003$ & $-0.049$, $0.058$ & $-0.047$ & $-0.084$, $-0.010$ & $-0.051$ & $-0.099$, $-0.005$ \\
\midrule
Pooled               & $-0.005$ & $-0.023$, $0.013$ & $-0.043$ & $-0.057$, $-0.029$ & $-0.040$ & $-0.054$, $-0.027$ \\
\bottomrule
\end{tabular}}
\end{table}

\begin{table}[!htbp]
\centering
\caption{\textbf{Accuracy in the development set (\%)} at a threshold of 0.5, and for A2 with one threshold per cancer cross-fitted in five folds within the development set.}
\label{tab:dev_accuracy}
\small
\begin{tabular}{lcccccc}
\toprule
Cancer & Majority & A1 & A2 & A3 & A4 & A2 + threshold \\
\midrule
Lung                 & 75.7 & 41.7 & 59.7 & 74.3 & 70.7 & 77.7 \\
Breast               & 93.5 & 69.2 & 85.4 & 89.5 & 90.1 & 93.3 \\
Brain (malignant)    & 83.9 & 41.7 & 77.2 & 79.4 & 82.8 & 86.7 \\
Stomach              & 80.9 & 29.9 & 66.2 & 79.0 & 82.2 & 87.9 \\
Bladder              & 85.1 & 55.7 & 84.1 & 82.1 & 83.1 & 85.6 \\
Biliary tract        & 75.8 & 25.8 & 59.1 & 75.0 & 72.0 & 78.0 \\
Liver                & 79.3 & 31.6 & 66.5 & 81.1 & 77.1 & 85.9 \\
Pancreas             & 72.6 & 29.6 & 41.1 & 69.7 & 71.6 & 78.6 \\
Colorectal           & 89.8 & 53.3 & 84.9 & 87.9 & 88.4 & 92.1 \\
Non-Hodgkin lymphoma & 91.7 & 51.9 & 89.6 & 91.4 & 92.3 & 92.3 \\
Prostate             & 91.5 & 67.2 & 89.7 & 90.9 & 90.5 & 92.2 \\
\bottomrule
\end{tabular}
\end{table}

\begin{table}[!htbp]
\centering
\caption{\textbf{Proportion of patients predicted to die at a threshold of 0.5 in the development set (\%),} by strategy and cancer, with the observed 90-day death rate for comparison.}
\label{tab:dev_predrate}
\small
\resizebox{\linewidth}{!}{%
\begin{tabular}{lrrrrrrrrrrr}
\toprule
 & Lung & Breast & Brain & Pancreas & Liver & Prostate & Colorectal & NHL & Biliary & Bladder & Stomach \\
\midrule
Observed death rate       & 24.3 & 6.5 & 16.1 & 27.4 & 20.7 & 8.5 & 10.2 & 8.3 & 24.2 & 14.9 & 19.1 \\
\midrule
A1 zero-shot, original wording & 82.7 & 35.9 & 72.2 & 97.8 & 89.1 & 40.5 & 56.5 & 54.8 & 98.5 & 57.2 & 89.2 \\
A2 zero-shot, matched wording  & 60.0 & 16.3 & 25.6 & 85.8 & 47.9 & 9.5 & 21.6 & 9.6 & 57.6 & 21.9 & 49.0 \\
A3 random                 & 28.0 & 8.4 & 17.8 & 39.9 & 19.9 & 6.0 & 14.0 & 4.3 & 26.5 & 15.9 & 24.8 \\
A4 similar                & 31.0 & 6.0 & 10.0 & 36.1 & 24.5 & 6.8 & 14.4 & 3.3 & 22.0 & 15.9 & 24.2 \\
A5 similar, shuffled      & 28.3 & 8.0 & 9.4 & 36.3 & 26.3 & 5.4 & 14.4 & 4.1 & 27.3 & 13.4 & 23.6 \\
R1 random, death-matched  & 38.7 & 12.5 & 12.8 & 39.2 & 31.9 & 10.5 & 19.3 & 5.3 & 31.1 & 17.4 & 27.4 \\
R2 similar (TF-IDF)       & 33.7 & 9.2 & 10.6 & 36.1 & 23.4 & 7.8 & 13.3 & 4.7 & 28.0 & 13.4 & 26.8 \\
S0 summary, zero-shot     & 61.7 & 15.5 & 26.1 & 77.6 & 49.7 & 15.7 & 22.8 & 14.7 & 52.3 & 23.9 & 43.3 \\
S3 summary, random        & 41.7 & 13.3 & 15.6 & 40.9 & 35.1 & 16.7 & 21.6 & 13.9 & 28.8 & 22.9 & 32.5 \\
S4 summary, similar       & 39.0 & 11.0 & 14.4 & 39.7 & 36.7 & 11.6 & 17.4 & 10.6 & 28.8 & 20.4 & 28.7 \\
\midrule
TF-IDF + LR               & 38.3 & 6.2 & 15.0 & 35.1 & 27.7 & 6.6 & 22.1 & 1.4 & 24.2 & 21.9 & 14.0 \\
Fusion (A2 + TF-IDF)      & 16.7 & 0.0 & 5.6 & 20.7 & 12.0 & 4.1 & 7.2 & 3.5 & 14.4 & 10.9 & 15.9 \\
\bottomrule
\end{tabular}}
\end{table}

% ---------------------------------------------------------------------
\clearpage
\suppnote{Comparison with the original study}{supp:original}

This note gives the comparison with the original study\citep{anil2026dynamic} that the Discussion summarises.

Our findings should not be read as contradicting the original report\citep{anil2026dynamic}, whose cohorts, endpoints and models differed from ours and whose data we could not re-examine (Extended Data Table~\ref{tab:reconcile}). One of its findings has a direct counterpart in our data: accuracy rises with dynamic examples. A second agrees only in part. In our data, shortening both the target note and the example notes to summaries raised AUROC. In the original study, in which the number of examples changed at the same time, summaries improved results for glioma, whereas for breast cancer with the 70B model two full-note examples gave a higher AUROC and accuracy (0.774 and 0.831) than five summarised examples (0.741 and 0.799). The findings on discrimination are mixed in the original study as well. With the 8B model, the gains in AUROC over zero-shot prompting were not significant, whereas with the 70B model the AUROC for glioma increased from 0.742 to 0.813. The main difference to be explained is therefore that ranking improved for glioma with Llama-3 70B but worsened in our cohorts. Differences between the two studies cannot establish which factor is responsible, but several are plausible.

The first is the language model. The two studies share a retrieval encoder but not an LLM, and models differ in which answer they tend to favour and in how strongly they follow examples. With Qwen3-32B, examples lowered AUROC by 0.01 to 0.02 instead of 0.03 to 0.04, so the size of the effect, and possibly its sign, may differ with Llama-3; running the random-example control with the original models would test the role of the model directly. The second is the balance of outcomes. About 39\% of the glioma patients in the original study survived beyond the horizon, which makes it likely that five neighbours display both outcomes. Our outcome was rarer (mortality of 7--27\% by cancer), and for 48\% of test records the five retrieved patients shared one outcome, nearly always survival, so that the examples offered no contrast between outcomes; the low death rate largely accounts for this, as five random patients were all survivors for 46\% of records. Library sizes, by contrast, were of a similar order of magnitude: the original study had about 400 training patients per fold and our test phase had 264 to 1{,}733 per cancer, with one library per cancer in both. The third is the relative informativeness of the target note and of its neighbours. A discharge summary may state the prognosis, the goals of care and the discharge destination outright, and our horizon was short, whereas survival months to years ahead, read from the notes of the first 180 days after diagnosis, may be harder to infer from the text alone. In the original study, the accuracy of the 8B model from full notes for glioma (65.7\%) was about the same as that of the majority outcome of the retrieved neighbours (65\%), whereas in our cohorts the AUROC of the note-alone prompt (0.86) far exceeded that of the neighbour vote (0.72); this comparison is itself limited because it sets accuracy against AUROC. The fourth is the outcome horizon. The original endpoints were survival beyond 14 months for glioma, close to the median survival of that cohort, and beyond 5 years for breast cancer. A model's general knowledge may indicate which features are favourable without indicating where a cohort-specific horizon divides the patients, and labelled cases from the same cohort can supply that information; predicting death within 90 days may depend less on it. This would fit the original observation that the rarer cancer with the poorer prognosis gained most. The fifth is the form of the examples. The best original configuration used summaries written to one template for each cancer, with fields for diagnosis, treatment and molecular markers that could be compared between target and example, whereas our full-note examples were free text of about 2{,}700 tokens each, and the target note made up about a sixth of the prompt.

These considerations lead to four hypotheses about when retrieved examples may improve prediction: (i) when the zero-shot model strongly favours one answer at the default threshold and examples shift that tendency; (ii) when the outcome is defined relative to the cohort, as with a long horizon near the median survival; (iii) when the examples are short, uniformly structured and contain the variables that drive prognosis; and (iv) when the cohort is homogeneous and its outcomes are balanced, so that five neighbours show both outcomes and their outcomes are about as informative as the target note. The original setting meets most of these conditions and ours meets few. Our threshold analyses bear directly on the first, although only for accuracy, and the summary-based strategies bear indirectly on the third, as the loss from similar examples shrank from 0.034 to 0.007 when notes were replaced by summaries that were neither templated per cancer nor reviewed. The second and fourth require another task, for example a one-year outcome in the same cohorts. A random-example control in the original cohorts would determine whether similarity contributed to the gains reported there.

\end{document}